\documentclass{article} %
\usepackage{amsmath, nccmath, bm}
\usepackage{enumitem}
\usepackage{graphicx, multicol}
\usepackage{graphicx,subcaption,ragged2e}
\usepackage[utf8]{inputenc}
\usepackage{listings}
\usepackage{xcolor}
\usepackage{tgtermes}
\usepackage{amssymb}
\usepackage{float}
\usepackage{comment}
\usepackage{caption}
\usepackage{soul}
\usepackage[margin=1in]{geometry}

\usepackage{capt-of}
\usepackage{tabu,booktabs}
\usepackage{authblk}
\usepackage{graphicx}
\usepackage[colorlinks,citecolor=blue,urlcolor=blue,bookmarks=false,hypertexnames=true,linkcolor=blue]{hyperref} 
\usepackage{ragged2e}
\usepackage{lineno}
\usepackage[mathscr]{euscript}

\usepackage{biblatex} 
\title{FO-PINN: A First-Order formulation for Physics-Informed Neural Networks}  

\author[1]{Rini Jasmine Gladstone}
\author[2]{Mohammad Amin Nabian}
\author[3]{N. Sukumar }
\author[4]{Ankit Srivastava}
\author[1]{Hadi~Meidani\thanks{Corresponding author : meidani@illinois.edu}}

\affil[1]{University of Illinois at Urbana-Champaign}
\affil[2]{NVIDIA}
\affil[3]{University of California Davis}
\affil[4]{Illinois Institute of Technology}

\date{}      

\begin{document}

\maketitle

\begin{abstract}
\justifying 
\noindent 
Physics-Informed Neural Networks (PINNs) are a class of deep learning neural networks that  learn the response of a physical system without any simulation data, and only by incorporating the governing partial differential equations (PDEs) in their loss function. While PINNs are successfully used for solving forward and inverse problems, their accuracy decreases significantly for parameterized systems. PINNs also have a soft implementation of boundary conditions resulting in boundary conditions not being exactly imposed everywhere on the boundary. With these challenges at hand, we present first-order physics-informed neural networks (FO-PINNs). These are PINNs that are trained using a first-order formulation of the PDE loss function. We show that, compared to standard PINNs,  FO-PINNs offer significantly higher accuracy in solving parameterized systems, and reduce time-per-iteration by removing the extra backpropagations needed to compute the second or higher-order derivatives. Additionally, FO-PINNs can enable exact imposition of boundary conditions using approximate distance functions, which pose challenges when applied on high-order PDEs. Through three examples, we demonstrate the advantages of FO-PINNs over standard PINNs in terms of accuracy and training speedup. 
\end{abstract}

\section{Introduction}
\label{introduction}
Numerical models are integral for scientific computing and have become an essential aspect in different fields such as civil and aerospace engineering, material science, physics and biology. Solving physics problems governed by partial differential equations (PDEs), often relies on high-fidelity solvers like finite element and finite volume techniques, which can become computationally expensive. An alternative to this are surrogate models which approximate solutions to the physical systems in a  faster and efficient manner. There have been many developments in deep learning algorithms and architectures, recently, to develop neural network surrogates for the same, which provides comparable accuracy as that of traditional solvers. They can be fully data-driven in their training or physics-based, where no simulation data used for training or a combination of both \cite{thuerey2021pbdl}. 

Widely used data driven methods for surrogate modeling are Convolutional Neural Networks (CNNs), DeepONet, Neural Operators and Graph Neural Networks (GNNs). CNNs use snapshots of observed data over a discretized domain to predict the physical solution. While such data-driven methods do not require a priori knowledge of the governing PDE, they are often limited to the specific domain discretization and cannot be easily generalized to other domain geometries. These CNN-based methods include PDE-inspired architectures, which  interpret image data as discretizations of multivariate functions and the output of image processing algorithms as solutions to certain PDEs \cite{ruthotto2020deep}, autoregressive dense encoder-decoder networks to model non-linear dynamical systems using smaller training dataset \cite{geneva2020modeling}, and  symbolic multi-layer neural networks which uses observed dynamic data with minor prior knowledge on the underlying mechanism  for its training \cite{long2019pde}. Recently, efforts in generalizing the performance of these architectures led to the development of neural operators and DeepONets, where the objective is to learn a continuous mapping between input and output of a PDE. This approach can be effective in accurately predicting the solution of several PDEs with the same form but different parameters.  Researchers have also built upon advances in GNNs to exploit the graph structure provided by the finite element mesh,  to glean local interactions among nodes for predicting system responses \cite{pfaff2020learning, fortunato2022multiscale}. Mesh-based GNNs encode in a connected graph the mesh information and the corresponding physical parameters such as loading and boundary conditions and model parameters \cite{gladstone2023gnn}.

Physics-Informed Neural Networks (PINNs)  \cite{cai2022physics, hennigh2021nvidia, https://doi.org/10.1111/mice.12685, raissi2019physics}, on the other hand, use physics-based deep learning framework to solve problems governed by partial differential equations (PDEs). In these approaches, the boundary and initial conditions of the system along with the governing PDEs are incorporated into the loss function.  The training of PINNs, which involves minimizing a purely physics-based loss function, doesn't require obtaining any training data,  which avoids the need to run finite element simulations. PINNs have been  used successfully in solving forward and inverse problems \cite{chen2020physics, GAO2022114502, hennigh2021nvidia, LOU2021110676, yang2021b}. The strength of PINNs lies in the flexibility of their application on a  variety of challenging PDEs \cite{cuomo2022scientific}, whereas traditional solvers typically require tailoring to the specifics of a particular PDE. In particular, this includes problems from computational physics that are notoriously hard to solve with classical numerical approaches due to strong nonlinearities, shocks, etc.  PINNs have been able to successfully solve nonlinear PDEs and problems such as Schrödinger, Burgers and Allen–Cahn equations.

While PINNs have good performance for strong-form PDES, they are not very effective for solving problems involving higher order PDEs \cite{luong2023deep}.  In PINNs, derivatives are calculated using automatic differentiation. It has been shown that as the order of the derivative increases the complexity in automatic differentiation increases and it becomes computationally expensive \cite{baydin2018automatic}. They also experience a decline in accuracy \cite{fresca2021comprehensive, mattey2022novel}, partially due to the sharp variations in the second and higher-order derivatives that are computed by automatic differentiation, which destabilize the network training. These problems get compounded and become more prominent when the responses of the physical systems are non-smooth and there are strong nonlinearities in the problems.  We also observe similar performance decline of PINNs for parametrized problems, as can be seen later in this paper.

Parameterized problems arise in many important application areas, including design optimization, uncertainty analysis, optimal control, and inverse problems and are a challenging problem to solve. The application of PINNs in this field has only been recently explored \cite{liu2023surrogate}. A parameterized problem means  solving the governing equations using PINNs for a range of input parameters. The input parameters can be PDE, geometric or boundary condition parameters.  This differs from the traditional finite element approach, where the governing equations are solved for one value of input parameters. Thus, while traditional method would require numerous simulations to cover a range of input parameters, a parameterized PINN can obtain the solutions in a single training. However, as the complexity of the problems increases with the number of parameters and higher order derivatives, the training cost significantly increases since the PDE residuals need to be evaluated on large numbers of collocation points in high-dimensional input spaces \cite{gao2021phygeonet}. In our previous experiments, we have observed that the accuracy of PINNs for parameterized systems quickly decline as the order of the PDE and the number of parameters of the problems increase. This could be because of the compounding errors in the higher order derivatives for high dimensional input space. 

The challenges with the higher order derivatives are also evident in the PINN algorithms for exact imposition of boundary conditions. Standard PINNs only offer soft imposition of boundary conditions compared to what is offered by other numerical methods such as finite element. Most of the existing PINN approaches enforce the boundary conditions by means of additional penalization terms that contribute to the loss function, which are multiplied by constant weighting factors. However, this leads to poor approximations and there have been methods to improve this. Instead of constant weights, some works propose the use of adaptive scaling parameters to balance the different terms in the loss functions. This still does not ensure hard implementation of boundary conditions. Using the theory of R-functions and approximate distance functions, \cite{SUKUMAR2022114333} introduced a generalized formulation for exact boundary condition imposition for PINNs. While this method has been successful in imposing Dirichlet, Neumann and Robin boundary conditions for complex geometries, it suffers from an exploding Laplacian issue for losses that involve second- or higher-order derivatives. Thus its application for a range of problems with higher order PDEs is severely limited.

Lastly, the presence of second and higher-order derivatives in the loss functions of PINNs causes issues when the widely used Automatic Mixed Precision (AMP) \cite{micikevicius2017mixed} is used for training. This is because second and higher-order derivatives require a different gradient scaling that is beyond the scope of AMP. Therefore, training of standard PINNs using AMP becomes unstable after a few iterations due to improper scaling. 

To address these challenges related to  high-order derivatives, we propose a novel scheme which redefines PINNs as an effective solution approach for higher-order problems. This is done by creating  a first-order formulation of the original problem. We show that this first-order formulation enables  PINNs to (1) solve the parameterized problems more accurately by smoothing out the sharp variations in the second and higher-order derivatives and by enabling exact boundary condition imposition, and (2) speed up training by reducing the number of required backpropagations and enabling the use of AMP for training. We organize the paper as follows: Section \ref{background} offers a brief background on PINN architecture and training as well as detailed information on exact boundary condition imposition using R-functions. Section \ref{method} provides an overview of the proposed FO-PINN approach and how they are used for boundary condition imposition. Section \ref{results} describes the numerical experiments and the results of three examples and the comparative performances between the proposed approach and high fidelity solvers. Finally, Section \ref{conclusion} concludes with a discussion on our contributions and directions for future work. 

\section{Background}
\label{background}

\subsection{Physics Informed Neural Networks (PINNs)}
\label{pinn}
Let us consider a physical system governed by the following generalized differential equation and boundary condition,

\begin{equation}
    \begin{aligned}
    \mathcal{N}(\bm{x};u(\bm{x})) = 0, \quad \bm{x} \in \mathcal{D}, \\
    \mathcal{B}(\bm{x};u(\bm{x})) = 0, \quad \bm{x} \in \partial \mathcal{D},
\label{eq_general}
    \end{aligned}
\end{equation}
where $\mathcal{N}(\cdot)$ is a differential operator and $\mathcal{B}(\cdot)$ is a boundary condition operator. Both the operators can consist of linear and nonlinear terms. In general, $\mathcal{N}(\cdot)$ consists of spatial derivatives of any order and  also temporal derivatives. Boundary conditions can be Dirichlet, Neumann or Robin and thus can include differential operators as well. $\bm{x}$ is the position vector in the  spatial domain $\mathcal{D}\subseteq \mathbb{R}^{D}$ bounded by the boundary $ \partial \mathcal{D}$, and $u(\bm{x})$ is the response of the physical system at a given location $\bm{x}$. 

In this work, we seek to approximate $u(\bm{x})$ using PINNs  given the governing equations  Eq.~\eqref{eq_general}. Specifically, consider a fully connected neural network, with a number of  hidden layers and a number of neurons in each layer.  The input to this network is the coordinates of a collocation point $\bm{x}$  in the computational domain and the output is the estimated response denoted by $\hat{u}(\bm{x};\bm{\theta})$, where $\bm{\theta}$ are the parameters of the neural network model, i.e. the weights and biases in the network.  The number of layers and neurons are the hyperparameters that are tuned during the training of the network. 

The PINN model  is trained by minimizing the loss function which consists of non-negative residuals of the governing PDE defined over the domain, $\mathcal{D}$ and that of the given boundary conditions defined on the boundary, $\partial \mathcal{D}$. These residuals are respectively given by

\begin{equation}\label{eqn:l2-redidual}
\begin{aligned}
r_\mathcal{N} (  \bm{\theta}  ) &=\int_{\mathcal{D} }( \mathcal{N} (\bm{x}; \bm{\theta} ) )^2  d\! \bm{x},   \\
r_\mathcal{B} (  \bm{\theta}  )&=\int_{ {\partial \mathcal{D}}} ( \mathcal{B} (  \bm{x}; \bm{\theta} ) )^2  d\!\bm{x}.   
\end{aligned}
\end{equation}

The optimal  parameters  of the PINN model, $\bm{\theta^*}$, can then be calculated according to 
\begin{equation} \label{loss}
\bm{\theta^*}=\underset{{ \bm{\theta} }}{\operatorname{argmin}}\, \underbrace{r_\mathcal{N}(  \bm{\theta} ) +\lambda r_\mathcal{B}  (  \bm{\theta}   )}_{J(  \bm{\theta} )},
\end{equation}
where, $\lambda$ controls the weight of boundary condition loss. This is a hyper parameter tuned during the training. The loss function is calculated using a set of collocation points during each iteration of training and it can be computed as 
\begin{equation} \label{loss-approximate}
\begin{split}
& J(\bm{\theta})  \approx  \frac{1}{m}\sum_{j \in M^{(i)}} J(\bm \theta; \bm x_j ) = \frac{1}{m}\sum_{j \in M^{(i)}} 
\bigg[ \left[\mathcal{N}\left( \bm{x}_j; \hat{u}(\bm{x}_j;\bm{\theta} ) \right) \right]^2  \\ &+
\lambda_1 \left[\mathcal{B} \left(\underbar{$\bm{x}$}_j ; \hat{u}(\underbar{$\bm{x}$}_j ;\bm{\theta} ) \right) \right]^2 \bigg],
\end{split}
\end{equation}
where $M^{(i)}$ is the set of indices of  selected collocation points at iteration $i$ with $|M^{(i)}|=m$, $J(\bm \theta; \bm x_j )$ is the per-sample loss evaluated at the $j$th collocation point, and $\{\underbar{$\bm{x}$}_j\}$ denotes the coordinates of the $j$-th boundary collocation point. The model parameters are updated according to
\begin{equation} \label{descent step}
\bm{\theta}^{(i+1)} = \bm{\theta}^{(i)} - \eta^{(i)} \nabla_{\bm{\theta}}{J}(\bm{\theta}^{(i)}),
\end{equation}
where $\nabla_{\bm{\theta}} J$ is calculated using backpropagation \cite{baydin2018automatic}.

\subsection{Exact BC Imposition using R Functions}
\subsubsection{Approximate Distance Functions (ADFs) using R functions}
\label{exact_bg}
Let $D \subset \mathbb{R}^d$ denote the computational domain with boundary $\partial D$. The exact distance to the boundary is the shortest distance between any point $\bm{x} \in  \mathbb{R}^d$ to the boundary $\partial D$ and therefore, is zero on $\partial D$. Computing the exact distance is a computationally expensive problem. The distance function may not be continuously differentiable at all points. Therefore, it becomes necessary to construct an ADF, $\phi (\bm{x})$, to overcome these challenges. If essential boundary conditions are imposed on the entire boundary $\partial D$, then the ADF must be zero on $\partial D$, positive in $D$, and its gradient must not vanish for any  $\bm{x} \in  \partial D$ .

Specifically, in 2D problems, let $\partial D \in \mathbb{R}^2$ be a complex boundary composed of $n$ line segments and curves, $D_i$, and let $\phi_i$ denote the ADF to each curve or line segment. Then the ADF of the entire geometry, $\phi$,  can be calculated as a function of the individual ADFs, $\phi_1, \phi_2, ..., \phi_n. $, as shown in Eq.~\eqref{eq:r_equi}. Following are the properties of the ADF function.

\begin{itemize}
    \item For any point $\bm{x}$ on $\partial D$, $\phi(\bm{x})=0$.
    \item $\phi(\bm{x})$ is normalized to the $m$-th order i.e. its derivative with respect to the unit inward normal vector is one and second to $m$-th order derivatives are zero for all the points on $\partial D$. This requirement ensures the closeness of ADF to the exact distance function. 
\end{itemize}

Let $F(\omega)$ be a function of a generalized variable, $\omega$. $F(\cdot)$ is considered to be a R-function if the sign of the value of $F$ is dependent only on $x$ (its positive if $x$ is positive and negative is $x$ is negative). The elementary properties of R-functions, including R-disjunction (union), R-conjunction (intersection) and R-negation, can be used for constructing a composite ADF, $\phi (\bm{x})$, to any arbitrarily complex boundary $\partial D$ , when ADFs  $\phi_i(\bm{x})$, to the partitions of $\partial D$ are known.

\subsubsection{ADF for Line Segments and Curves}

Let us consider a line segment of length $L$ with end points $\bm{x}_1 \equiv (x_1,y_1)$ and $\bm{x}_2 \equiv (x_2,y_2)$, midpoint $\bm{x}_c=(\frac{x_1+x_2}{2},\frac{y_1+y_2}{2})$ and length $L = ||\bm{x}_2-\bm{x}_1||$. Then ADF for the line segment, $\phi(\bm{x})$ can be calculated as follows.
\begin{equation}
\begin{aligned}
f(\bm{x}) &= \frac{(x-x_1)(y_2-y_1)-(y-y_1)(y_2-y_1)}{L} \\
t(\bm{x}) &= \frac{1}{L}[L^2-||\bm{x}-\bm{x}_c||^2] \\
\Phi &= \sqrt{t^2 + f^4} \\
\phi(\bm{x}) &= \sqrt{f^2+(\frac{\Phi - t}{2})^2} 
\label{eq:1}
\end{aligned}
\end{equation}

Similarly ADF for a circle of radius $R$ and center located at $\bm{x}_c \equiv (x_c, y_c)$ is given by

\begin{equation}
\begin{aligned}
\phi(\bm{x}) = \frac{R^2-(\bm{x}-\bm{x}_c).(\bm{x}-\bm{x}_c)}{2R} 
\label{eq:ad_circle}
\end{aligned}
\end{equation}

The derivation of the ADF calculation of line segments and curves can be found in \cite{SUKUMAR2022114333}. Once the ADFs, $\phi_i(\bm{x})$ to all the partitions of $\partial D$ are calculated, we can calculate the ADF to $\partial D$ using the R-equivalence operation. When $\partial D$ is composed of $n$ pieces,$\partial D_i$, then ADF, $\phi$, that is normalized up to order m is given by

\begin{equation}
\begin{aligned}
\phi(\bm{x})=\frac{1}{\sqrt[m]{\frac{1}{(\phi_1(\bm{x}))^m}+\frac{1}{(\phi_2(\bm{x}))^m}+...+\frac{1}{(\phi_n(\bm{x}))^m}}}\\
\label{eq:r_equi}
\end{aligned}
\end{equation}

\begin{figure*}[t]
\begin{subfigure}[t]{0.32\textwidth} 
\vbox{
\vspace*{0.2em}%
\centering{
  \includegraphics[width=\textwidth]{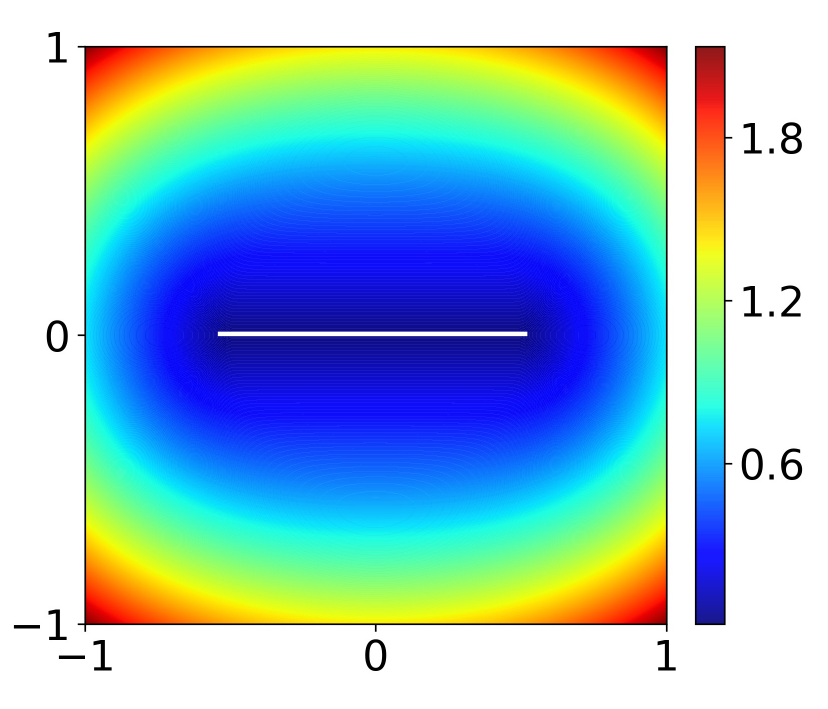}
         \caption{ADF for a line segment}
         \label{fig:adf_line}
}%
\vspace*{0.2em}
}%
\end{subfigure}%
\hfill
\begin{subfigure}[t]{0.32\textwidth} 
\vbox{
\vspace*{0.2em}%
\centering{
\includegraphics[width=\textwidth]{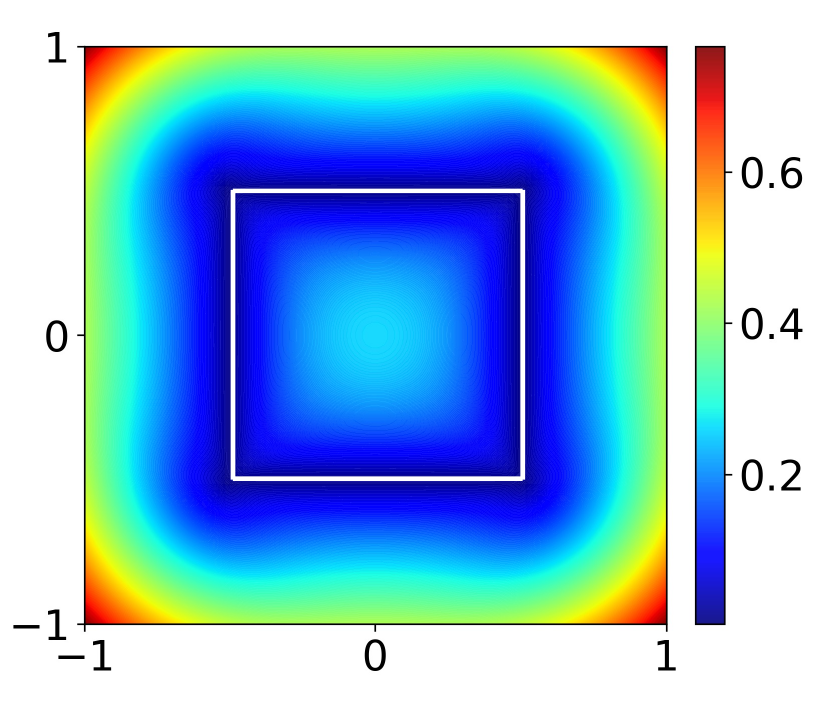}
\caption{ADF for a line segment}
\label{fig:adf_square}
}
\vspace*{0.2em}
}
\end{subfigure}%
\hfill
\begin{subfigure}[t]{0.32\textwidth} 
\vbox{
\vspace*{0.2em}%
\centering{
\includegraphics[width=\textwidth]{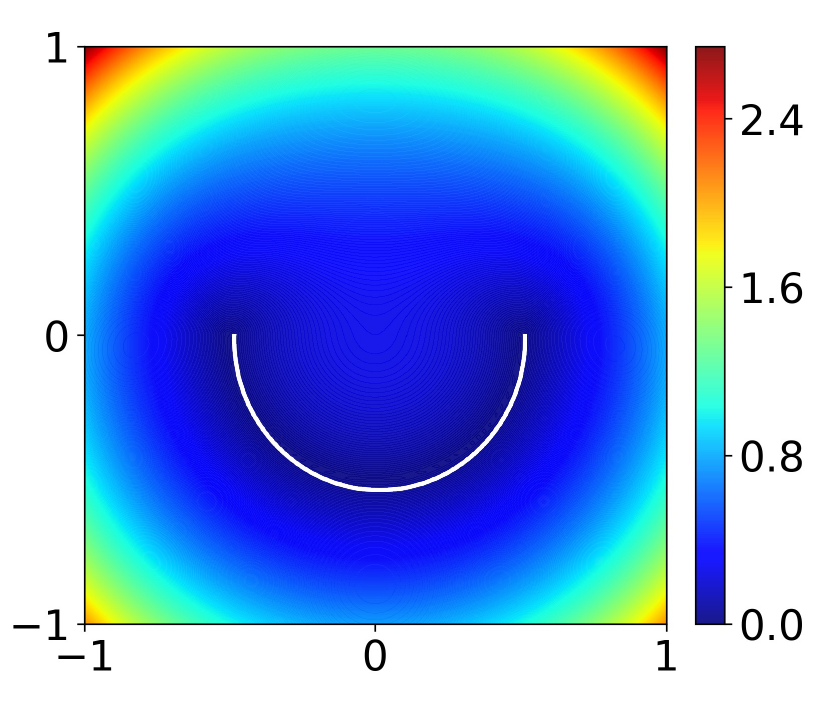}
         \caption{ADF for a semi circle}
         \label{fig:adf_semicircle}
}
\vspace*{0.2em}
}
\end{subfigure}%
     \caption{The ADFs for various shapes calculated using Eq.~\eqref{eq:1}, ~\eqref{eq:ad_circle}, ~\eqref{eq:r_equi}. The shape for which ADF is calculated is highlighted in white. ADF is closer to zero near the shape and it increases as you go further away from the shape.}
        \label{fig:adfs}
\end{figure*}

The images in Fig. \ref{fig:adfs} show the ADFs for line segment, circle/arc and polygon. The ADFs were calculated using the above equations. For the square (polygon), the R-equivalence operation defined in Eq.~\eqref{eq:r_equi} was used to calculate the final ADF from the ADFs of the individual line segments.

\subsubsection{Exact Boundary Condition Imposition}

For Dirichlet boundary condition, if $u=g$ is prescribed on $\partial D$, then the solution structure to impose exact boundary condition is given by

\begin{equation}
\begin{aligned}
u = g + \phi \tilde{u}, \\
\label{eq:3}
\end{aligned}
\end{equation}
where  $\tilde{u}$ is the approximate solution, in this case, the PINN solution.

Similarly the solution structure for Neumann boundary condition, $\frac{\partial u}{\partial n}=h$ on $\partial D$ is given by

\begin{equation}
\begin{aligned}
u = [1+\phi D_1^\phi](\tilde{u}_1)-\phi h + \phi^2 \tilde{u}_2, \\
\label{eq:4}
\end{aligned}
\end{equation}
where $\tilde{u}_1$ and $\tilde{u}_2$ are approximate functions obtained through PINN and $D_1^\phi(·)$ is a differential operator that acts in the outward normal direction to the boundary. This solution structure can be modified as given below, for Robin boundary condition, $\frac{\partial u}{\partial n} + cu = h$ on $\partial D$.

\begin{equation}
\begin{aligned}
u = [1+\phi(c+D_1^\phi)](\tilde{u}_1)-\phi h + \phi^2 \tilde{u}_2, \\
\label{eq:5}
\end{aligned}
\end{equation}

When different inhomogeneous essential boundary conditions are imposed on distinct subsets of $\partial D$, we can use transfinite interpolation to calculate $g$ function, the boundary condition function for the entire boundary $\partial D$. On using the singular inverse-distance based Shepard weight function, we can write the transfinite interpolant as

\begin{equation}
\begin{aligned}
& g(\bm{x}) =  \sum_{i=1}^{M} w_i(\bm{x})g_i(\bm{x}) \\
& w_i(\bm{x}) = \frac{\phi_i^{-\mu_i}}{\sum_{j=1}^{M}\phi_j^{-\mu_j}} = \frac{\prod_{j=1;j \neq i}^{M} \phi_j^{-\mu_j}}{\sum_{k=1}^{M}\prod_{j=1;j \neq k}^{M} \phi_j^{-\mu_j}}, \\
\label{eq:6}
\end{aligned}
\end{equation}
where weights $w_i$ add up to one, and interpolates $g_i$ on the set $\partial D_i$. $\mu_i \geq 1$ is a constant controlling the nature of interpolation. This boundary value function, $g(\bm{x})$, can be used in Eq.~\eqref{eq:3} to calculate the final solution with the exact imposition of BC.

This approach, however, has certain challenges especially when solving second and higher order problems. ADFs using R function are not normalized at the joining points of lines and curves. A detailed explanation of this can be found in \cite{biswas2004approximate}. This results in the second and higher order derivatives of the solutions becoming undefined at the vertices/ joining points at the boundary.

\section{Method}
\label{method}
\subsection{First-order Physics-Informed Neural Networks (FO-PINNs)}

Without loss of generality, consider the following second order linear partial differential equation in two variables with constant coefficients and defined on the domain $D$,

\begin{equation}
    \begin{aligned}
    &a \frac{\partial^2 u}{\partial x^2} + b \frac{\partial^2 u}{\partial x \partial y} + c \frac{\partial^2 u}{\partial y^2} + d \frac{\partial u}{\partial x} + e \frac{\partial u}{\partial y} + f u = g(x,y), {x,y} \in \mathcal{D}, \\
    &\mathcal{B}(u,{x,y}) = 0, {x,y} \in \partial \mathcal{D},
\label{eq_general_pde}
    \end{aligned}
\end{equation}
where $B$ is the boundary conditions of the system, which can be Dirichlet, Neumann or mixed and $u$ is the response of the system. For the ease of demonstration, we consider a time independent problem. However, this is easily applied to a time dependent problem with initial conditions as well. In standard PINNs, the output of the neural network consists of the dependent variables in the PDE that is being solved, which is ths response of the system, $u$. The higher order derivatives of the response, required in the PDE loss calculation are computed directly using these outputs by automatic differentiation. 

In FO-PINNs, for solving a PDE (of order $d$), in addition to the dependent variables, neural network output also consists of the derivatives (of order  up to  $d-1$) of the dependent variables. The second and higher-order PDEs are reformulated as a series of first-order PDEs with additional compatibility equations to ensure the compatibility between the predicted and exact derivatives. These compatibility equations are incorporated as additional terms in the loss function. With this formulation, automatic differentiation is used only to compute first-order derivatives of the network outputs w.r.t. inputs, and this significantly reduces the number of required backpropagations compared to standard PINNs. 

For Eq.~\eqref{eq_general_pde}, in FO-PINNs, the first-order derivatives $\frac{\partial u}{\partial x}$,  $\frac{\partial u}{\partial y}$ are defined as new (output) variables $u_x$ and $u_y$ respectively. Thus, Eq.~\eqref{eq_general_pde} is reformulated to produce the following governing equation on the new output variables:

\begin{equation}
    \begin{aligned}
    &a \frac{\partial u_x}{x} + b  \frac{\partial u_x}{y}+ c \frac{\partial u_y}{y} + d \frac{\partial u}{x} + e \frac{\partial u}{y} + f u = g(x,y),
\label{eq_firstorder_general}
    \end{aligned}
\end{equation}
with the following compatibility equations:
\begin{equation}
    \begin{aligned}
    &u_x = \frac{\partial u}{\partial x}, 
    u_y = \frac{\partial u}{\partial y}.
    \label{eq_firstorder_compatibility_general}
    \end{aligned}
\end{equation}

\begin{figure*}[t]
\begin{subfigure}[t]{0.97\textwidth} 
\vbox{
\vspace*{0.2em}%
\centering{
  \includegraphics[width=\textwidth]{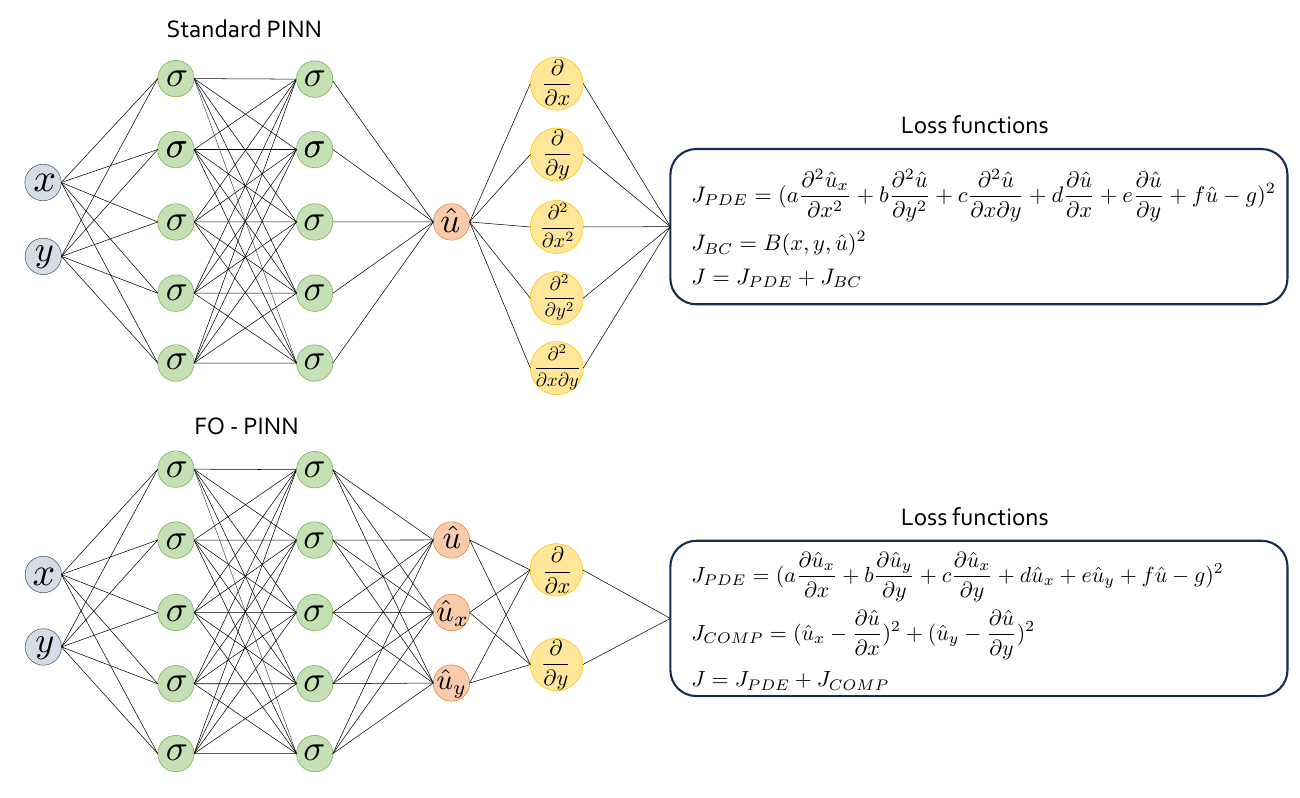}
}%
\vspace*{0.2em}
}%
\end{subfigure}%
     \caption{Overview of the FO-PINN architecture. The architecture at the top is of the standard PINN, with the output being the response of the physical system. The loss function consists of PDE and BC losses. For second order PDEs, we need to calculate first and second order derivatives for PDE loss calculation. The bottom one is the FO-PINN architecture with the outputs being the response of the physical system and the first order derivatives of the response. The loss function consists of the PDE loss and compatibility loss to ensure compatibility between the predicted and actual derivatives. There is no BC loss in this case, as we impose exact BC using R-functions.}
        \label{fig:methodology}
\end{figure*}

Thus, the output of a FO-PINN model includes ${u}_x$ and ${u}_y$ as well as ${u}$. In addition to the PDE and boundary condition losses, the loss function also consists of compatibility loss terms according to Eq.~\eqref{eq_firstorder_compatibility_general}. The first order spatial derivatives, i.e. $\frac{\partial {u}}{\partial x}$ and $\frac{\partial {u}}{\partial y}$ are calculated using automatic differentiation. An overview of the FO-PINN approach is shown in Fig. \ref{fig:methodology}.

\subsection{Exact Imposition of BCs for PINNs using ADFs}
\label{challenges_exact}
Exact imposition of boundary conditions (BC) in PINNs is challenging and non-trivial. Most of the purposed approaches are limited to square domains and suffer from convergence issues due to over-constrained solutions. A viable approach for complex geometries has been introduced in \cite{SUKUMAR2022114333}. It consists of calculating the ADF to the boundaries using the theory of R-functions, and formulating a boundary-condition and geometry-aware solution ansatz. 

Let $D \subset \mathbb{R}^d$ denote the computational domain with boundary $\partial D$. Let $\phi(\bm{x})$ be the ADF such that $\phi(\bm{x})=0$ for any point $\bm{x}$ on $\partial D$. For Dirichlet boundary condition, if $u=g$ is prescribed on $\partial D$, then the solution ansatz is given by $u_{sol} = g + \phi u_{net}$, where $u_{sol}$ is the approximate solution, and $u_{net}$ is the neural network output. The calculation of ADFs for various geometries and solution ansatz for Neumann, Robin and mixed boundary conditions are given in \cite{SUKUMAR2022114333} and explained briefly in section \ref{exact_bg}. This approach, however, suffers from an exploding Laplacian issue at the corner of the geometries when the governing  PDEs consist of second or higher-order derivatives. By bypassing second and higher-order derivative computations, FO-PINNs enable the use of this exact boundary condition imposition approach for improved solution accuracy. Furthermore, we improve the stability and time-to-convergence of the FO-PINNs, which are trained with exact BC imposition, by normalizing and non-dimensionalizing the PDE losses. This is done by normalizing the domains to be in unit scale and making sure all the parameters are dimensionless by appropriate scaling.

\section{Results}
\label{results}
In this section, we present three examples which demonstrate the advantages of FO-PINNs over standard PINNs. In section \ref{exact_bc_results}, we show the accuracy gain of FO-PINNs, with exact BC imposition using ADFs, when applied to the Helmholtz problem. In section \ref{section_annular_ring}, we demonstrate the performance of of FO-PINNs for an annular ring problem governed by Navier-Stokes equations. In section \ref{section_cylinder}, we compare the performance of FO-PINNs over the standard PINN solvers for parameterized systems governed by the Navier-Stokes equations. Both standard PINNs and FO-PINNs trained in the following examples are fully connected networks with 6 layers, 512 neurons per layer and Swish activation function. The networks were trained using the Adam optimizer. We used a single V100 GPU for training the models. Both the models were trained using Modulus by NVIDIA. 

\subsection{Exact BC Imposition with FO-PINNs}
\label{exact_bc_results}

For the first example, consider the following Helmholtz equation:
\begin{equation}
    \begin{aligned}
    k^2u+\frac{\partial^2u}{\partial x^2}+\frac{\partial^2u}{\partial y^2}+\frac{\partial^2u}{\partial z^2} = f,
    \label{eq_helmholtz}
    \end{aligned}
\end{equation}
where $k$ and $f$ are the wave number and source term, respectively. We consider a square domain of length 2 units with $u=0$ for the entire boundary. The wave number, $k$ is set to $1$ and the source term, $f$ is set to zero. 

For a standard PINN, the output of the network is the response, $u$. For FO-PINN, three additional variables, $u_x$, $u_y$ and $u_z$, are added to the output, where these represent the first-order derivatives, $\frac{\partial u}{\partial x}$,  $\frac{\partial u}{\partial y}$, and $\frac{\partial u}{\partial z}$ respectively. The Eq.~\eqref{eq_helmholtz} is reformulated to produce the following governing equation on the new output variables:

\begin{equation}
    \begin{aligned}
    &k^2u+\frac{\partial u_x}{\partial x}+\frac{\partial u_y}{\partial y}+\frac{\partial u_z}{\partial z} = f,
\label{eq_firstorder}
    \end{aligned}
\end{equation}
with the following compatibility equations:
\begin{equation}
    \begin{aligned}
    &u_x = \frac{\partial u}{\partial x}, 
    u_y = \frac{\partial u}{\partial y}, 
    u_z = \frac{\partial u}{\partial z}. 
    \label{eq_firstorder_compatibility}
    \end{aligned}
\end{equation}

The exact BC is imposed using the solution from FO-PINNs by defining the solution ansatz as described in section \ref{challenges_exact}. $usol = g + \phi unet$, where $unet$ is the approximate solution from FO-PINN. Here, $g=0$ as $u=0$ for the entire boundary. The ADF function $\phi$ is calculated using Eq.~\eqref{eq:r_equi}, where there are 4 individual ADFs, $\phi_1$, $\phi_2$, $\phi_3$ and $\phi_4$ corresponding to the four boundary lines of the square domain. The domain is normalized to be of unit length and unit area. The individual ADFs are calculated using Eq.~\eqref{eq:1}, where $L=1$ and the co-ordinates of the four vertices are $(0,0), (0,1), (1,0), (1,1)$. 

\begin{figure*}[t]
\begin{subfigure}[t]{0.45\textwidth} 
\vbox{
\vspace*{0.2em}%
\centering{
         \includegraphics[width=\textwidth]{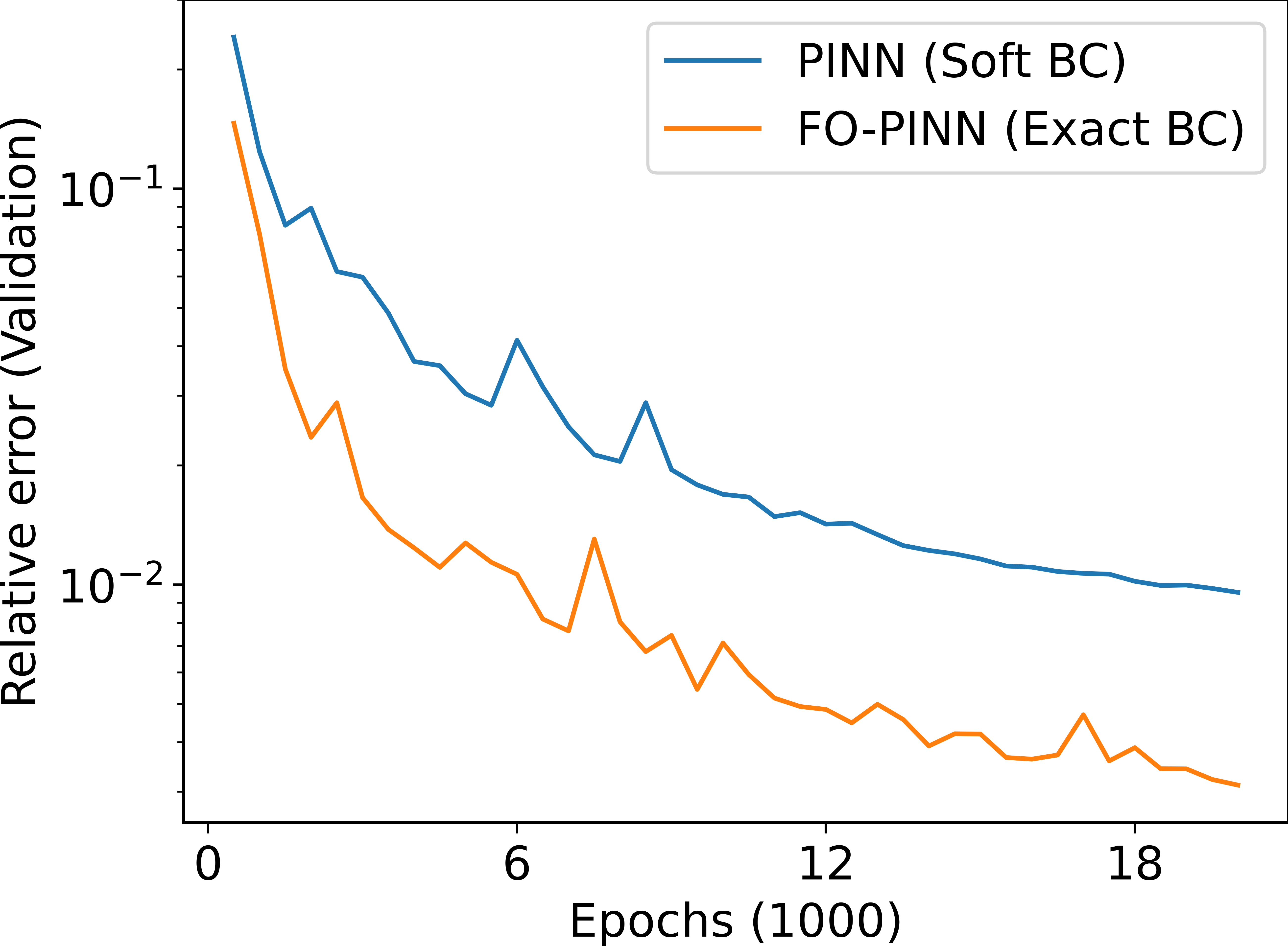}
         \caption{Error plot for predicted $u$}
         \label{fig:helmholtz_val}
}%
\vspace*{0.2em}
}%
\end{subfigure}%
\hfill
\begin{subfigure}[t]{0.45\textwidth} 
\vbox{
\vspace*{0.2em}%
\centering{
         \includegraphics[width=\textwidth]{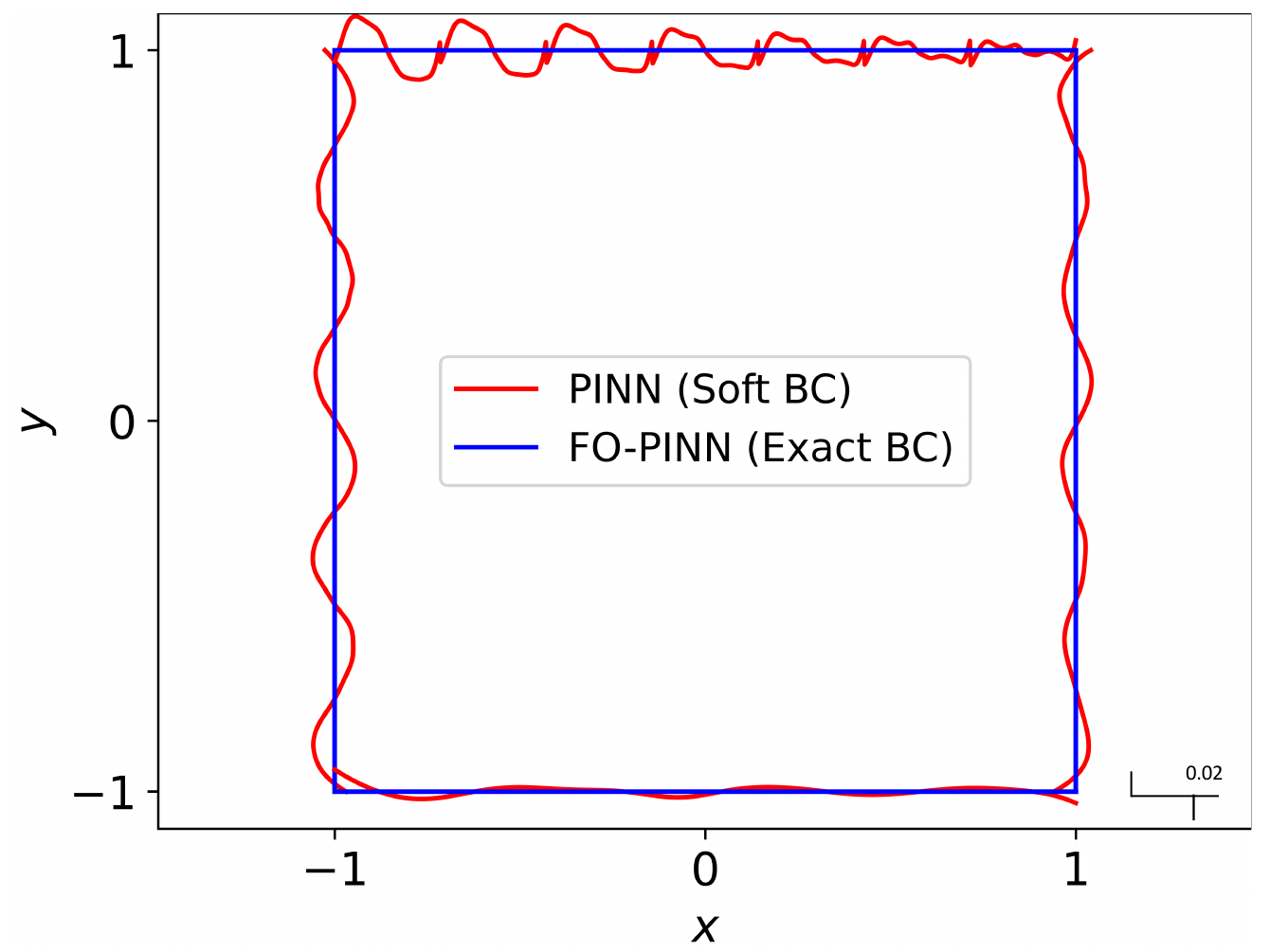}
         \caption{Predicted ${u}$ at the boundary}
         \label{fig:helmholtz_boundary}
}
\vspace*{0.2em}
}
\end{subfigure}%
     \caption{Results for Helmholtz equation on a square domain. The relative error is computed using the analytical solution for this example. From Fig. (a) we can observe that FO-PINN performs better than standard PINN in terms of relative error inside the domain, with the added benefit of exact BC imposition. Figure (b) shows the predicted $\hat{u}$ at the boundaries where the BC of $u=0$ is implemented. FO-PINN is able to exactly impose the BC using R-functions in this case. }
        \label{fig:helmholtz}
\end{figure*}

Figure \ref{fig:helmholtz} shows the performance of FO-PINN in predicting the response of the system accurately. From Fig. \ref{fig:helmholtz_val} we observe that the relative error for the validation data of FO-PINN with exact BC imposition is about one order of magnitude lower than that of PINN with soft BC imposition. The relative error is computed by comparing the predicted response to the analytical solution for this example. The analytical solution is given by $u(x) = -(-\pi^2sin(\pi x)sin(4 \pi y)-16\pi^2sin(\pi x)sin(4 \pi y)+sin(\pi x)sin(4 \pi y))$. It is to be noted that standard PINNs cannot implement exact BC using ADFs due to the challenges described in section \ref{challenges_exact}, and are thus trained with a soft BC imposition. Figure \ref{fig:helmholtz_boundary} shows the predicted ${u}$ at the boundary for both the models. While FO-PINN ensures that ${u}$ is exactly $0$ for the entire boundary, PINN with soft BC predicts the boundary values with an error of magnitude $10^{-2}$. 

\begin{figure*}[t]
\begin{subfigure}[t]{0.45\textwidth} 
\vbox{
\vspace*{0.2em}%
\centering{
\includegraphics[width=\textwidth]{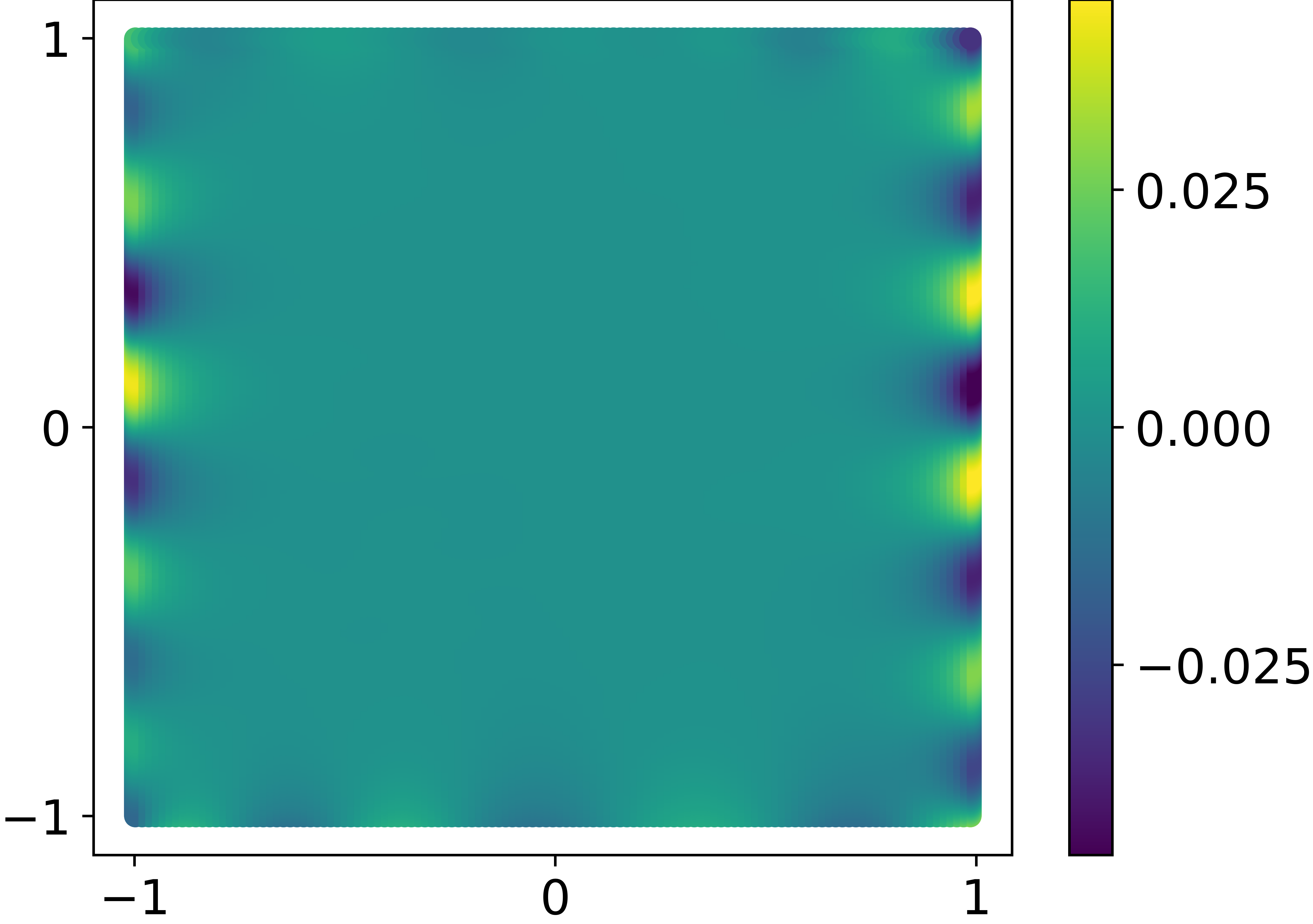}
         \caption{Error on $u$ prediction - Standard PINN}
         \label{fig:helmholtz_standard_diff}
}%
\vspace*{0.2em}
}%
\end{subfigure}%
\hfill
\begin{subfigure}[t]{0.45\textwidth} 
\vbox{
\vspace*{0.2em}%
\centering{
         \includegraphics[width=\textwidth]{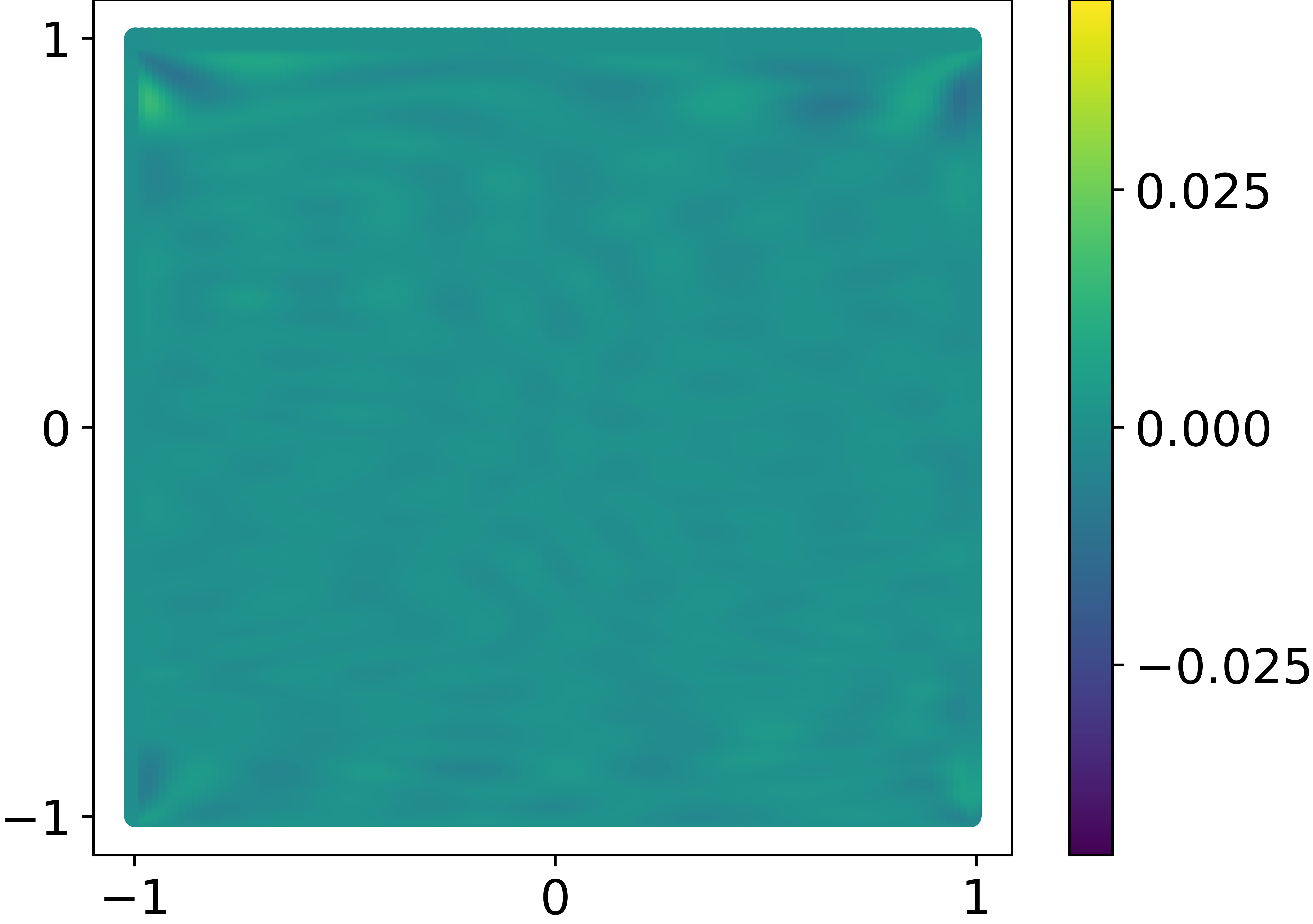}
         \caption{Error on $u$ prediction -  FO-PINN}
         \label{fig:helmholtz_adf_diff}
}
\vspace*{0.2em}
}
\end{subfigure}%
     \caption{The error distribution of the predicted response over the entire domain for the Helmholtz problem. The error is computed as the absolute difference between analytical solution for this example and the predicted output.}
        \label{fig:helmholtz_pred_diff}
\end{figure*}

The training of FO-PINNs is ~2.2x faster than that of PINNs (0.010 seconds per iteration for FO-PINN compared to 0.022 seconds per iteration for PINN with the same network architecture). This can be attributed to the reduced number of backpropagation steps required for the training of FO-PINNs. Additionally, we observed that training of FO-PINNs using AMP can achieve 1.33x speedup (0.0075 seconds per iteration) compared to FO-PINNs without the use of AMP. These two performance gains combined makes the training of FO-PINNs significantly faster compared to the training of PINNs (i.e., 2.9x speedup in this example). 

\subsection{FO-PINNs for Steady State Navier-Stokes Problem}
\label{section_annular_ring}

\begin{figure}[ht]
\centering\includegraphics[width=0.8\linewidth]{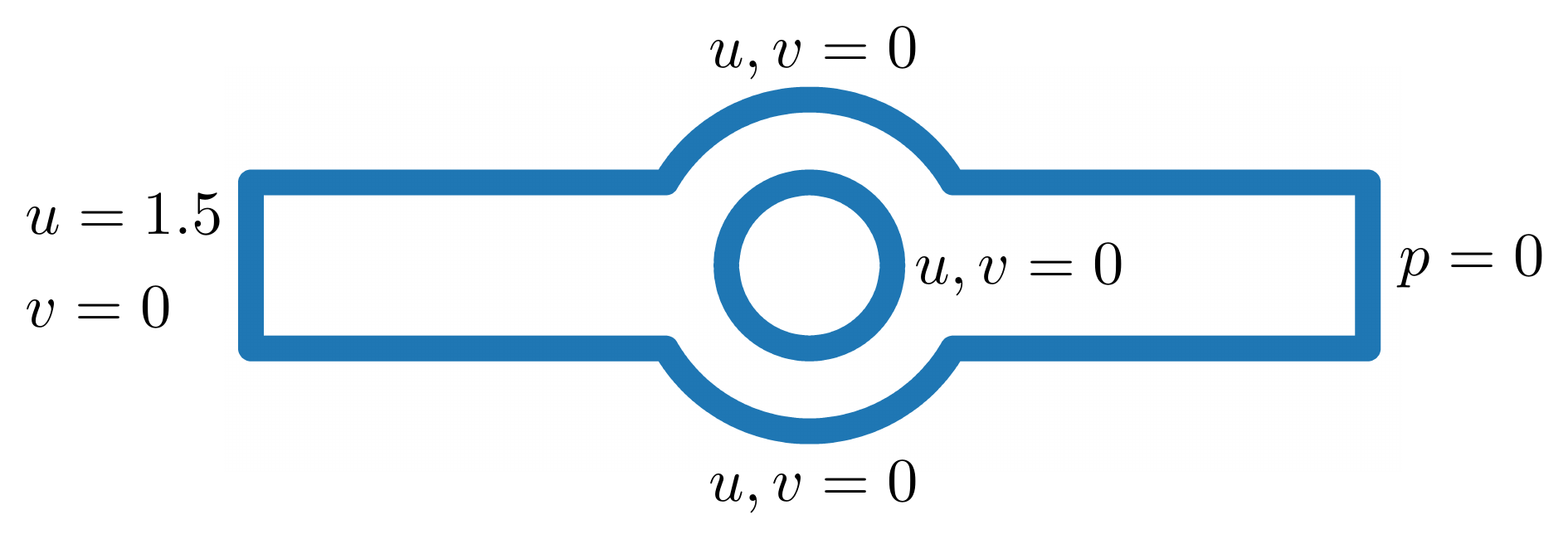}
\caption{Annular ring domain with prescribed boundary conditions.}
\label{fig:annular_ring_domain}
\end{figure}

For the second example, we consider an annular ring domain with Dirichlet boundary conditions governed by steady state Navier-Stokes equations. For this we consider annular ring with length 13.464 units and width 2 units, with two concentric circles of radii 2 and 1 unit respectively at the centre. The dimensions of the domain are scaled for the annular ring to be of unit length. The responses of the system are $u$, $v$ and $p$, which are the velocities in $x$ and $y$ directions and pressure respectively. The boundary conditions at inlet are $u=inlet_{vel}, v=0$ and that of outlet is $p=0$. The no-slip boundary conditions are $u=0$ and $v=0$. For this example, we have set $inlet_{vel}=1.5$. The domain with the boundary conditions is shown in Fig. \ref{fig:annular_ring_domain}. 

The governing equations defining the flow through the annular ring are given by continuity and momentum equations in $x$ and $y$ directions respectively as the following:

\begin{equation}
    \begin{aligned}
    \frac{\partial u}{\partial x}+\frac{\partial v}{\partial y} &= 0,\\
    \rho(u\frac{\partial u}{\partial x}+v\frac{\partial u}{\partial y}) &= -\frac{\partial p}{\partial x} + \nu(\frac{\partial^2 u}{\partial x^2}+\frac{\partial^2 u}{\partial y^2})\\
    \rho(u\frac{\partial v}{\partial x}+v\frac{\partial v}{\partial y}) &= -\frac{\partial p}{\partial y} + \nu(\frac{\partial^2 v}{\partial x^2}+\frac{\partial^2 v}{\partial y^2}).\\
    \label{eq_ns}
    \end{aligned}
\end{equation}

The kinematic viscosity, $\nu$ and fluid density, $\rho$ are set to be $0.01$ and $1$ respectively. For a standard PINN, the outputs of the network are the predicted responses $\hat{u}$, $\hat{v}$ and $\hat{p}$. However, for FO-PINN, apart from these outputs, additionally, we predict the first order spatial derivatives of $u$ and $v$ as well, which are denoted,  $\hat{u}_x$,  $\hat{u}_y$, $\hat{v}_x$ and  $\hat{v}_y$. Therefore, additional loss terms are introduced to ensure compatibility over these spatial derivatives. The compatibility loss terms for this problem are the following. 
\begin{equation}
    \begin{aligned}
    &\hat{u}_x = \frac{\partial \hat{u}}{\partial x}, 
    \hat{u}_y = \frac{\partial \hat{u}}{\partial y},\\ 
    &\hat{v}_x = \frac{\partial \hat{v}}{\partial x}, 
    \hat{v}_y = \frac{\partial \hat{v}}{\partial y},\\ 
    \label{eq_firstorder_compatibility_ns}
    \end{aligned}
\end{equation}
The momentum and continuity equations in Eq \ref{eq_ns} are reformulated to be in the first-order form as follows. 
\begin{equation}
    \begin{aligned}
     u_x+v_y &= 0,\\
    \rho(u u_x+v  u_y) &= -\frac{\partial p}{\partial x} + \nu(\frac{\partial u_x}{\partial x}+\frac{\partial u_y}{\partial y}),\\
    \rho(u  v_x+v  v_y) &= -\frac{\partial p}{\partial y} + \nu(\frac{\partial v_x}{\partial x}+\frac{\partial v_y}{\partial y}).\\
    \label{eq_ns_firstorder}
    \end{aligned}
\end{equation}

\begin{figure}[ht]
\centering\includegraphics[width=0.7\linewidth]{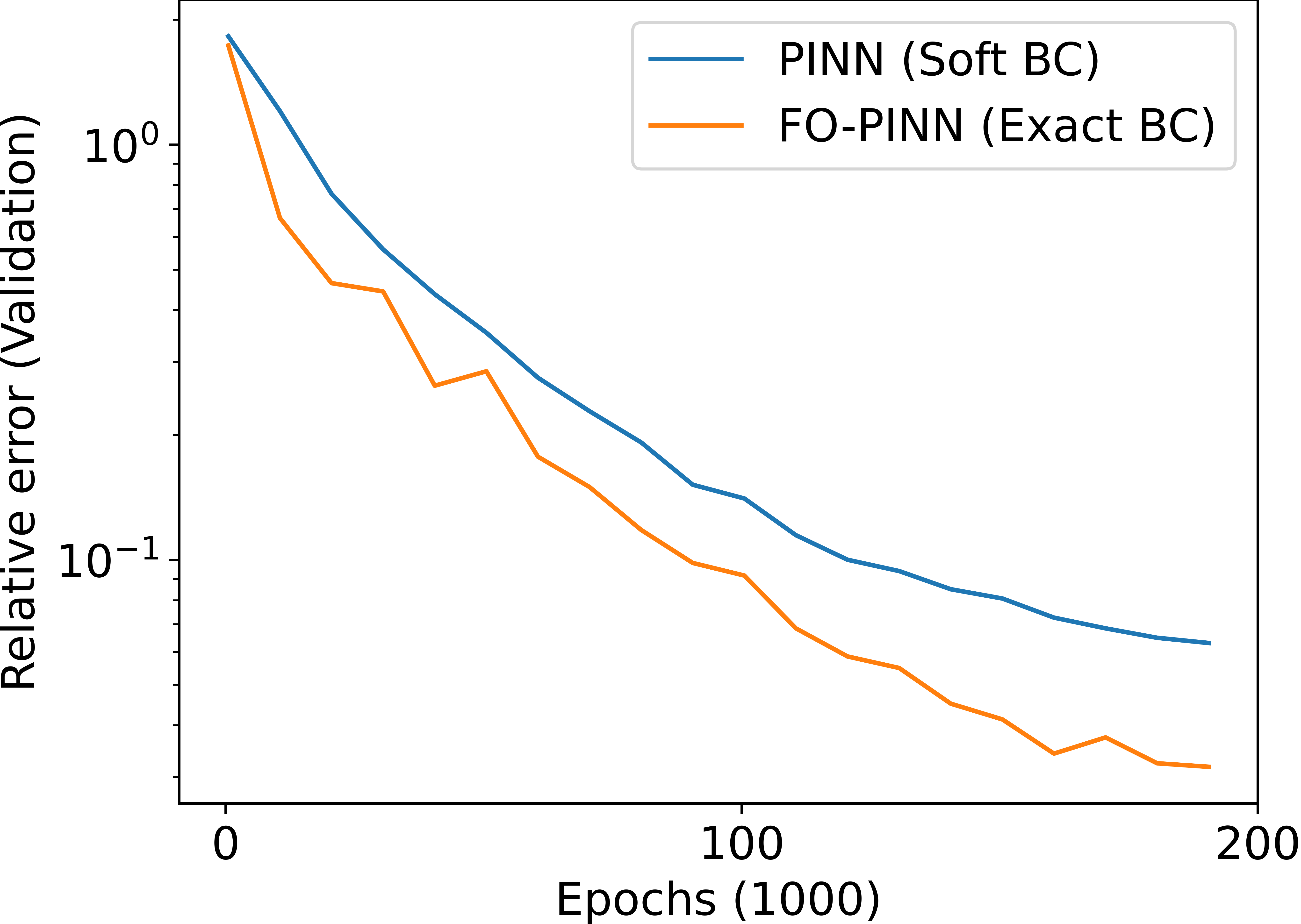}
\caption{Relative (validation) error for the prediction of $p$ on the validation dataset.}
\label{fig:annular_p_error}
\end{figure}

\begin{figure*}[t]
     \centering
     \begin{subfigure}[t]{0.3\textwidth}
\vbox{
\vspace*{0.2em}%
\centering{
         \includegraphics[width=\textwidth]{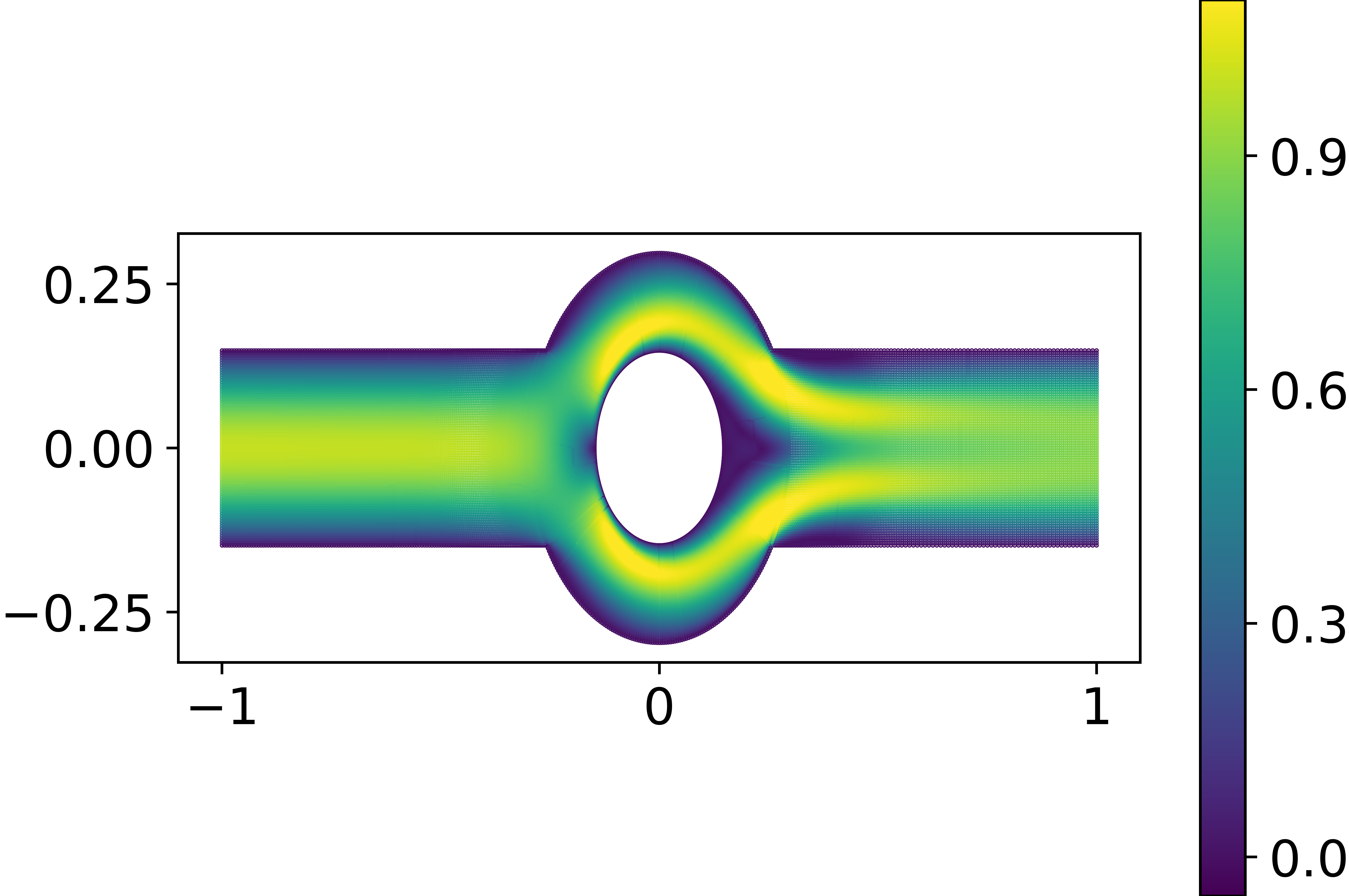}
         \caption{Velocity - Exact}
         \label{fig:annular_uv_exact}
         }%
\vspace*{0.2em}
}%
     \end{subfigure}
     \hfill
     \begin{subfigure}[t]{0.3\textwidth}
\vbox{
\vspace*{0.2em}%
\centering{
         \includegraphics[width=\textwidth]{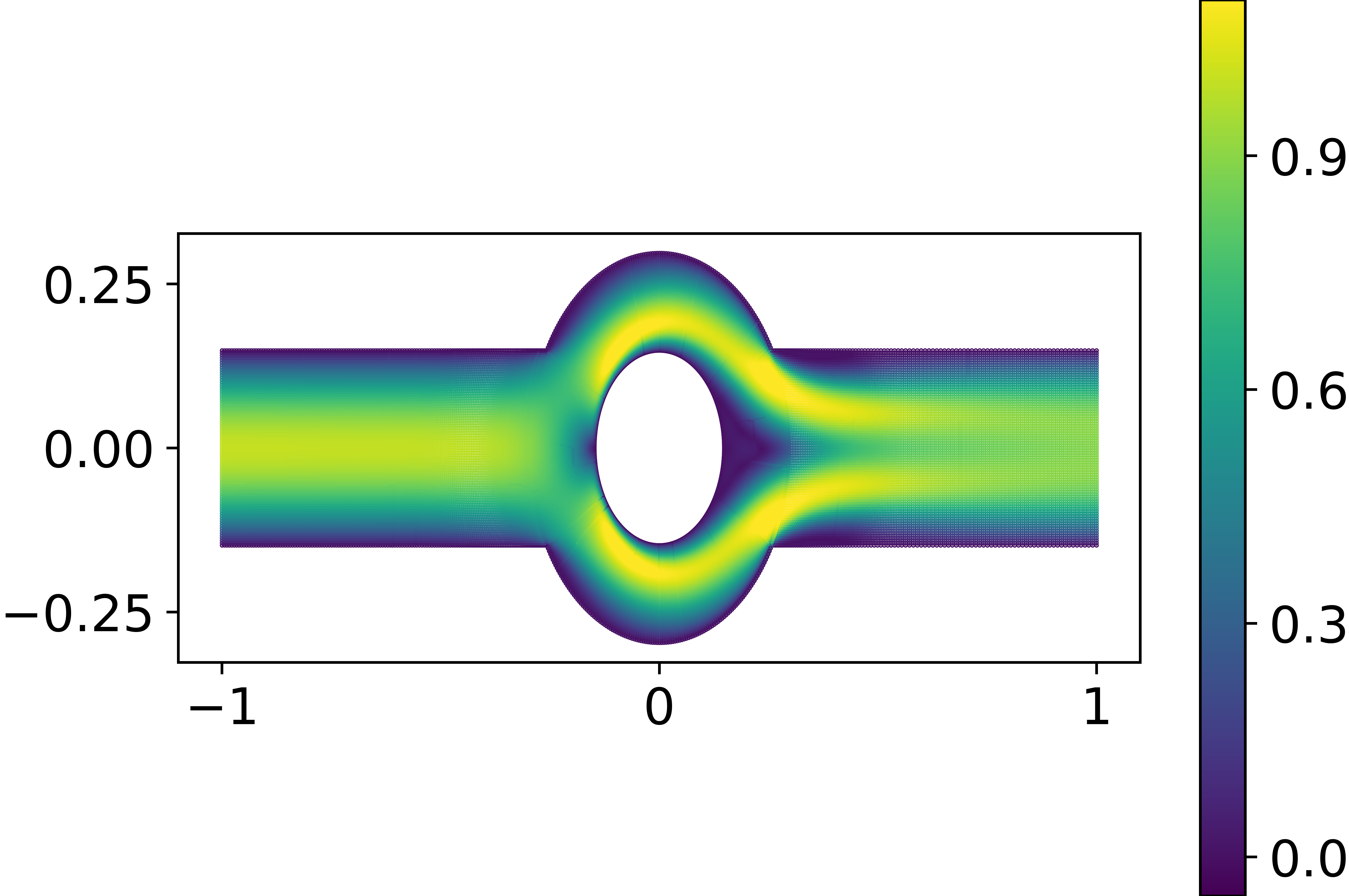}
         \caption{Velocity - PINN}
         \label{fig:annular_uv_pred_pinn}
         }%
\vspace*{0.2em}
}%
     \end{subfigure}
     \hfill
     \begin{subfigure}[t]{0.3\textwidth}
\vbox{
\vspace*{0.2em}%
\centering{
         \includegraphics[width=\textwidth]{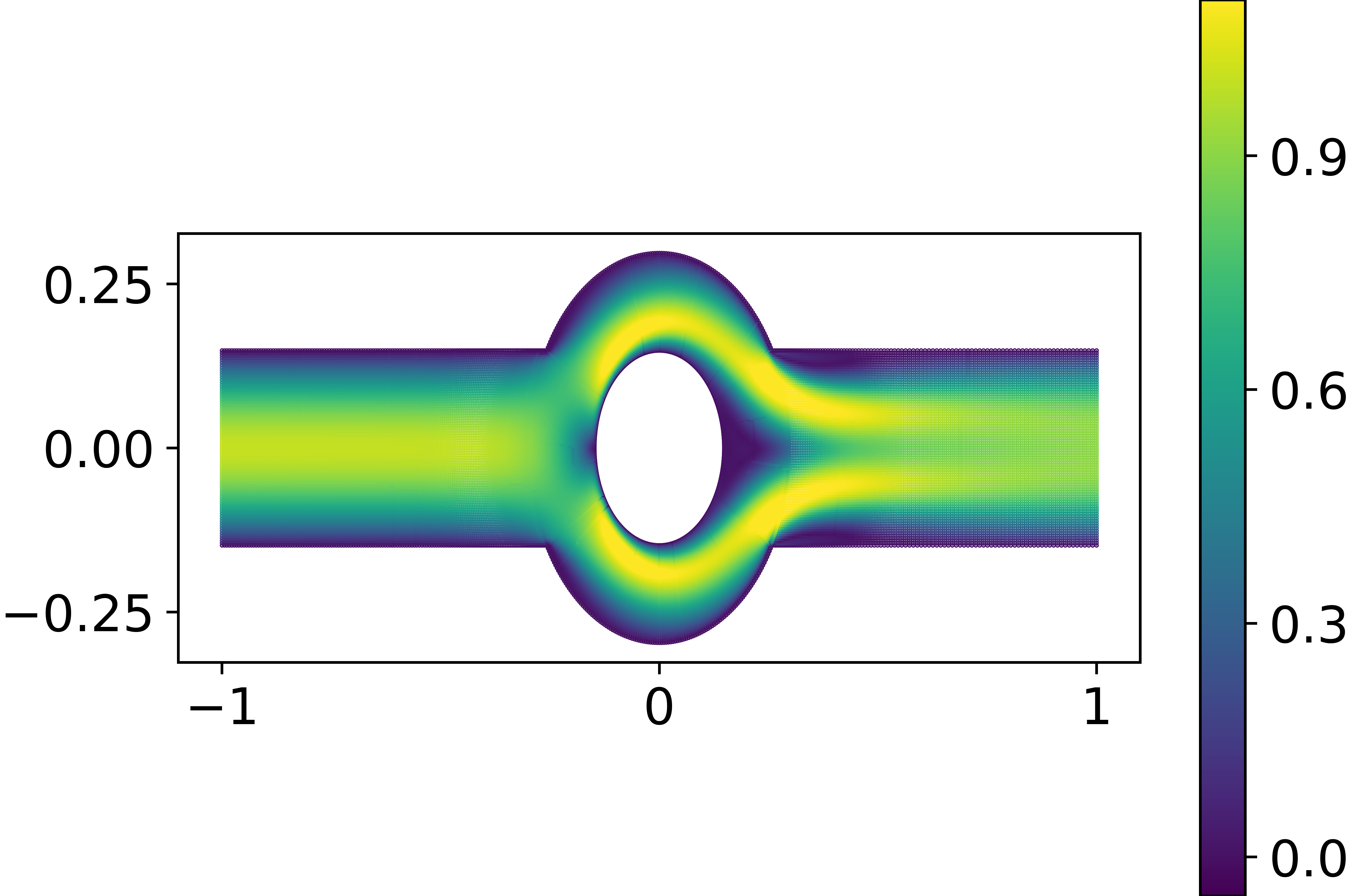}
         \caption{Velocity - FO-PINN}
         \label{fig:annular_uv_pred_fopinn}
         }%
\vspace*{0.2em}
}%
     \end{subfigure}
         \bigskip
     \begin{subfigure}[t]{0.3\textwidth}
\vbox{
\vspace*{0.2em}%
\centering{
         \includegraphics[width=\textwidth]{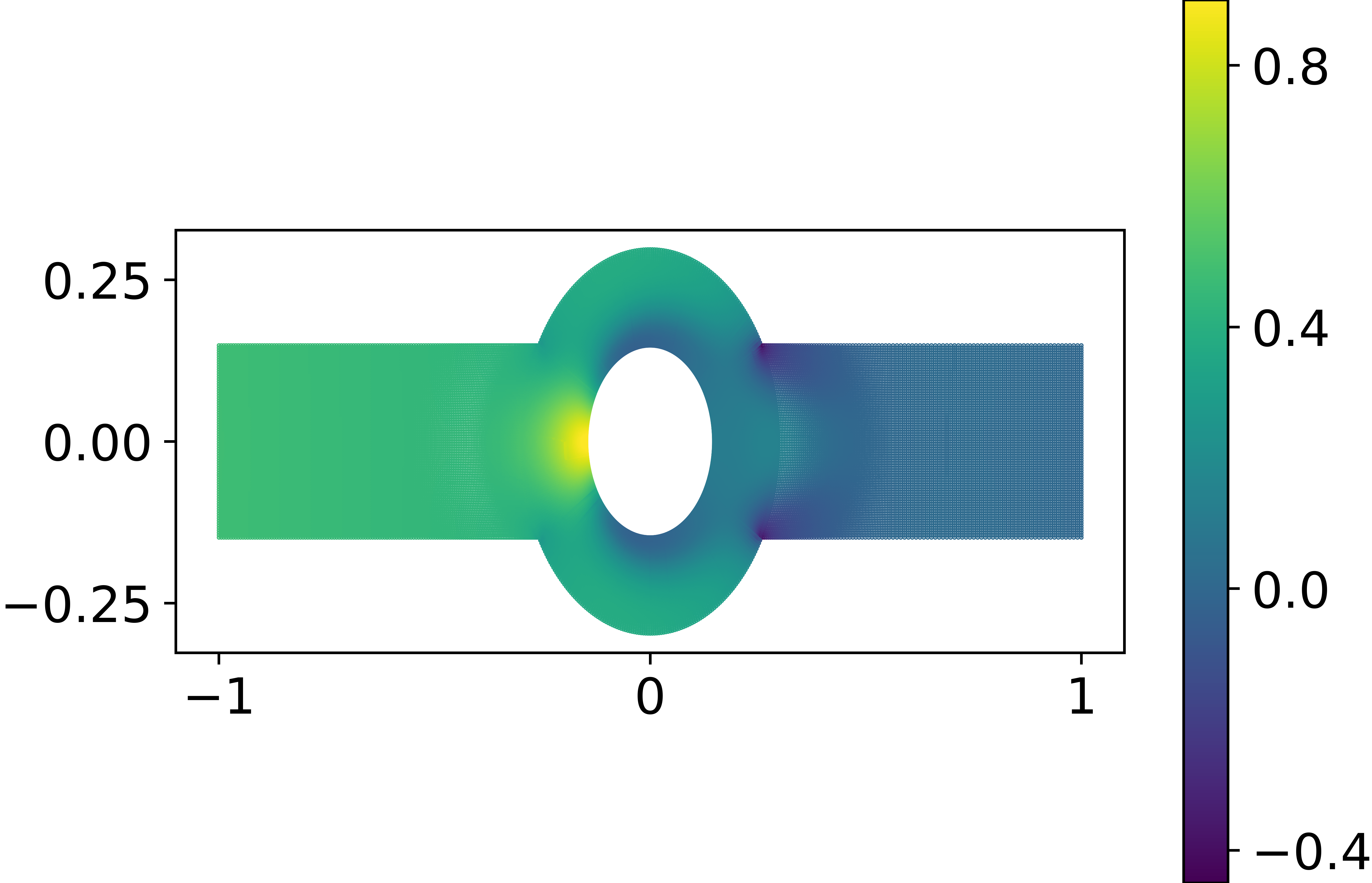}
         \caption{Pressure - Exact}
         \label{fig:annular_p_pred_exact}
         }%
\vspace*{0.2em}
}%
     \end{subfigure}
       \hfill
     \begin{subfigure}[t]{0.3\textwidth}
\vbox{
\vspace*{0.2em}%
\centering{
         \includegraphics[width=\textwidth]{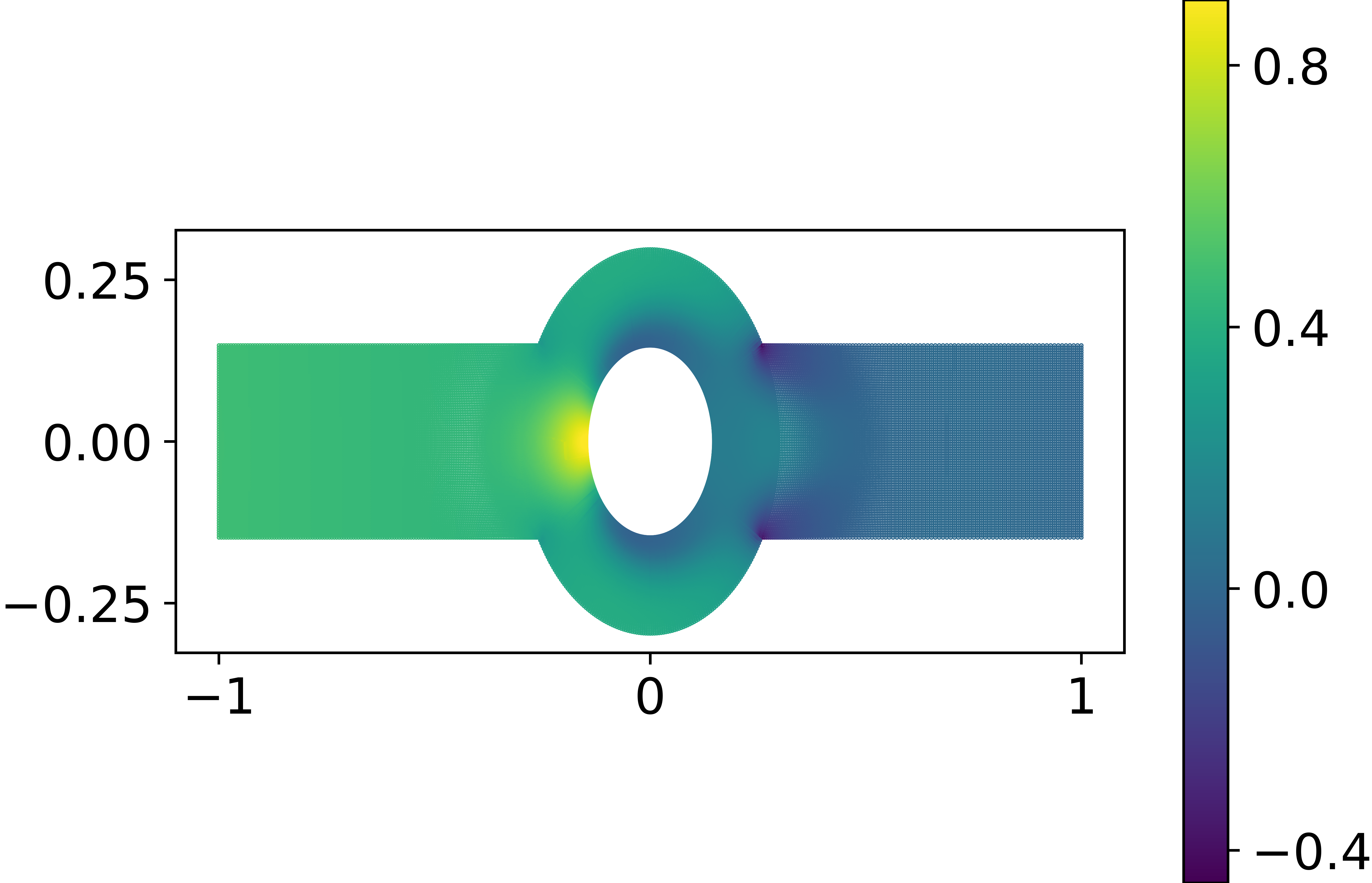}
         \caption{Pressure - PINN}
         \label{fig:annular_p_pred_pinn}
         }%
\vspace*{0.2em}
}%
     \end{subfigure}
     \hfill
     \begin{subfigure}[t]{0.3\textwidth}
\vbox{
\vspace*{0.2em}%
\centering{
         \includegraphics[width=\textwidth]{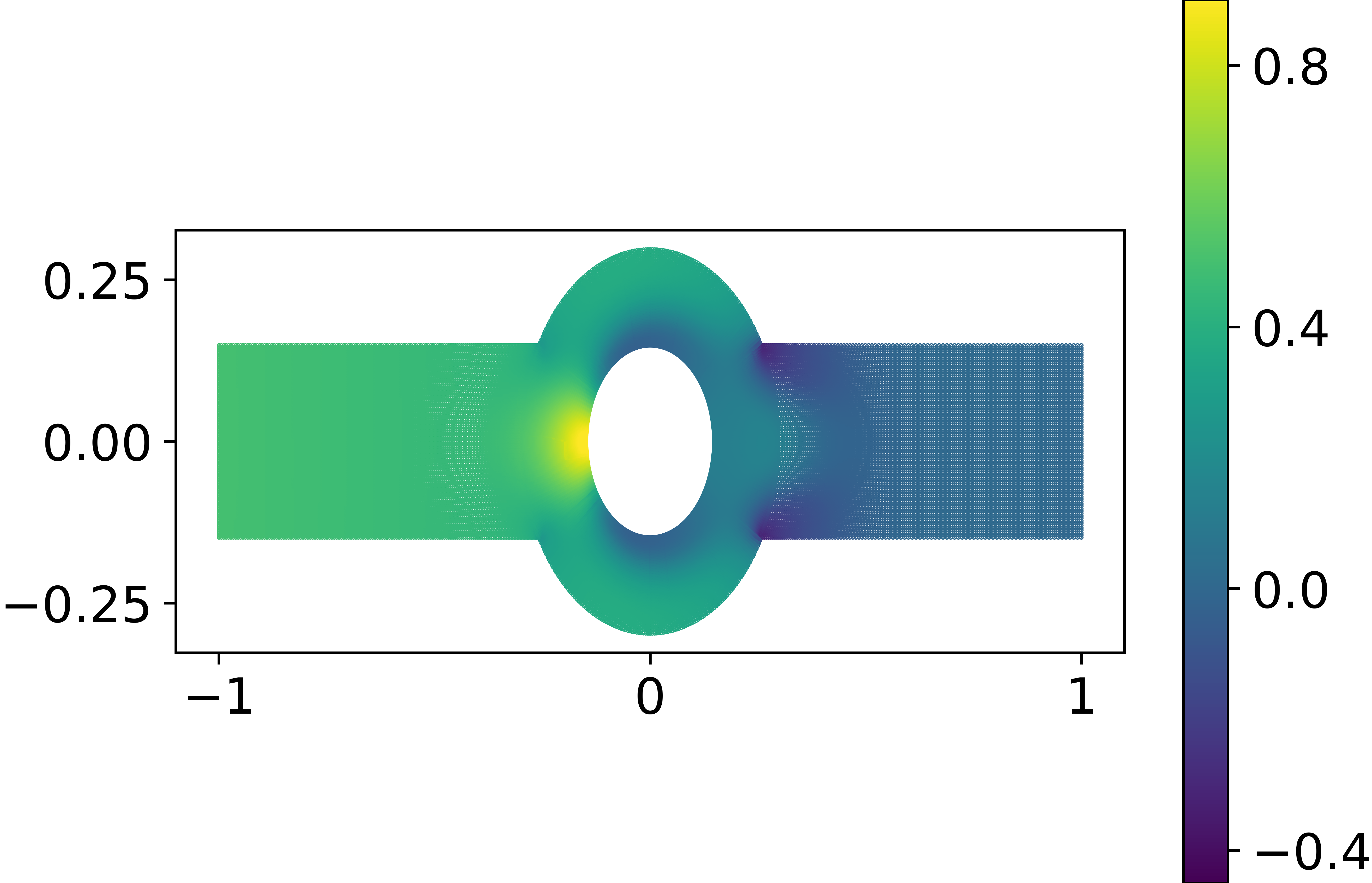}
         \caption{Pressure - FO-PINN}
         \label{fig:annular_p_pred_fo}
         }%
\vspace*{0.2em}
}%
     \end{subfigure}
    \caption{Predicted flow on the annular ring governed by Navier-Stokes equations using PINN and FO-PINN compared with the simulation results from OpenFOAM (Exact). Figures (a) through (c) show the exact velocity magnitude and the predicted values from the two networks. Similarly, Figures (d) through (f) show the exact and predicted pressure.}
    \label{fig:annular_ring_predicted_flow }
\end{figure*}

In addition to first-order formulation, we also impose exact boundary conditions using R-functions, similar to the previous example. Since we have a boundary with inhomogeneous dirichlet boundary conditions, we use Eq.~\eqref{eq:r_equi} and Eq.~\eqref{eq:6} together to formulate the ADF for this example. We use high-fidelity solutions from OpenFOAM simulations to compare the predictions from the neural network surrogates and assess their performance. Figure \ref{fig:annular_p_error} shows the validation error for the prediction of $p$ for standard PINN which has soft implementation of BCs and FO-PINN with exact BC imposition and Fig. \ref{fig:annular_ring_predicted_flow } shows the predicted flow on the annular ring domain by both the networks (standard PINN and FO-PINN) compared with that of the simulation results (which is considered as the reference solution here). From these figures, we can observe that FO-PINN has a better prediction accuracy compared to that of PINN, with the added benefit of imposition of the exact boundary conditions, making sure the error at the boundaries to be zero. For standard PINN, the boundary error is to the order of $10^{-2}$.   

\subsection{FO-PINNs for Parameterized Systems}
\label{section_cylinder}

While standard PINNs perform well for a problem with fixed PDE parameters and specific boundary and initial conditions, its performance declines very quickly for parameterized systems. This includes parameterized PDEs, geometries as well as boundary conditions. For the second example, we study the performance of FO-PINNs on parameterized systems and compare it with that of standard PINNs. 

\begin{figure}
\centering\includegraphics[width=0.7\linewidth]{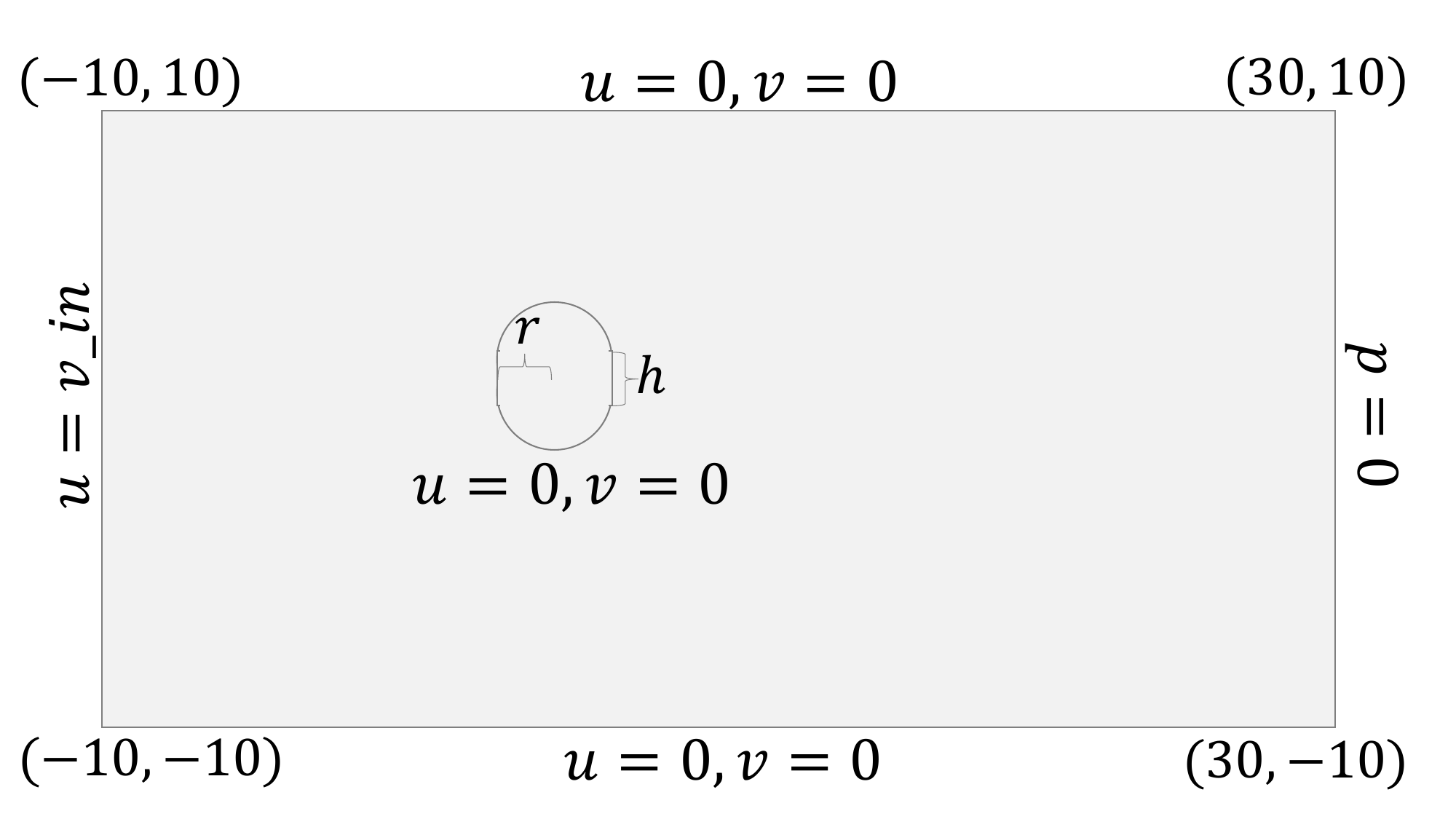}
\caption{Domain of the channel with cylinder slice and prescribed boundary conditions.}
\label{fig:channel_domain}
\end{figure}

For this, we consider a channel, of unit length and height of 0.5 units, with a cylinder slice located inside. We predict the flow through the channel, which is governed by the Navier-Stokes equations. The channel is parameterized with the radius of the cylinder slice $r$ is set to be in the range $[0.05,0.2]$ and the location of the center of the cylinder $(a,b)$ to be in the range $[-0.2,0.2]$ and $[-0.1,0.1]$ respectively. The channel has an inlet velocity, $v_{in}$, which is also considered as a parameter in the parameterized system and set to be in the range $[0.05,0.15]$. The outlet of the channel has zero pressure ($p=0$) and no slip boundaries have $u=0$ and $v=0$. The domain with the boundary conditions are shown in Fig. \ref{fig:channel_domain}. 

\begin{figure}
\centering\includegraphics[width=0.7\linewidth]{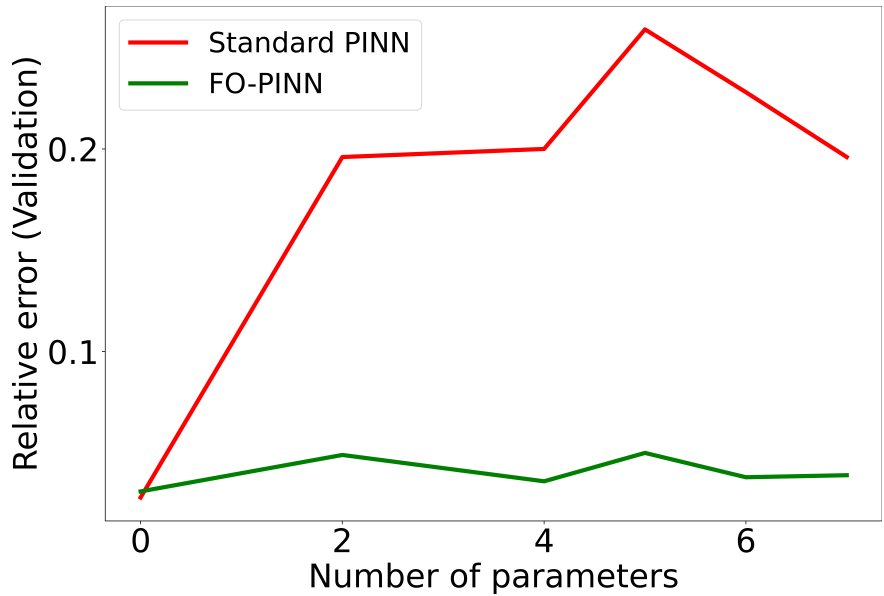}
\caption{Relative (validation) error trend for the prediction of $u$ as the number of parameters increase. The first two parameters are the location coordinates of the center of the cylinder, the next two parameters are the radius and height of the cylinder respectively, the fifth parameter is the inlet velocity, $v_{in}$. The last two parameters are kinematic viscosity, $\nu$ and fluid density, $\rho$ respectively. All the models are trained for 200,000 epochs.}
\label{fig:param_trend}
\end{figure}


The governing equations defining the flow through the channel is given by the continuity and momentum equations defined in Eq.~\eqref{eq_ns}. We consider steady state conditions and hence, the time components are ignored for the analysis. The kinematic viscosity, $\nu$ and fluid density, $\rho$ are also parameterized for the training of the network and are set to be in the range $[0.01,0.03]$ and $[0.8,1.2]$ respectively. Thus, our goal is to train neural networks solvers for a parameterized system  with a total of seven parameters - four geometric parameters, $a$, $b$, $r$ and $h$, two PDE parameters $\nu$ and $\rho$,  and one BC parameter, $v_{in}$.

\begin{figure*}[h]
\begin{subfigure}[t]{0.32\textwidth} 
\vbox{
\vspace*{0.2em}%
\centering{
\includegraphics[width=\textwidth]{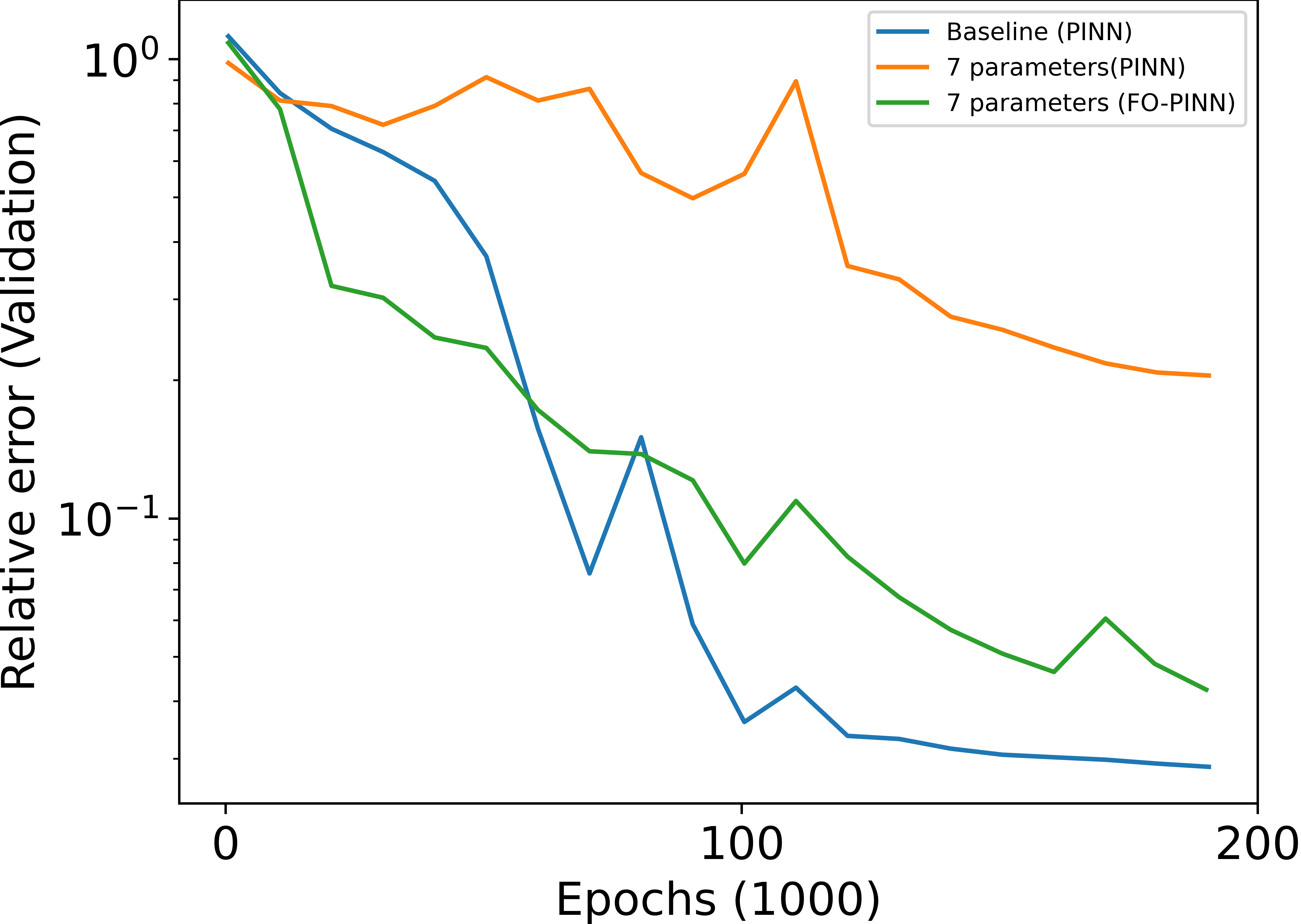}
         \caption{$u$}
         \label{fig:eu}
}%
\vspace*{0.2em}
}%
\end{subfigure}%
\hfill
\begin{subfigure}[t]{0.32\textwidth} 
\vbox{
\vspace*{0.2em}%
\centering{
          \includegraphics[width=\textwidth]{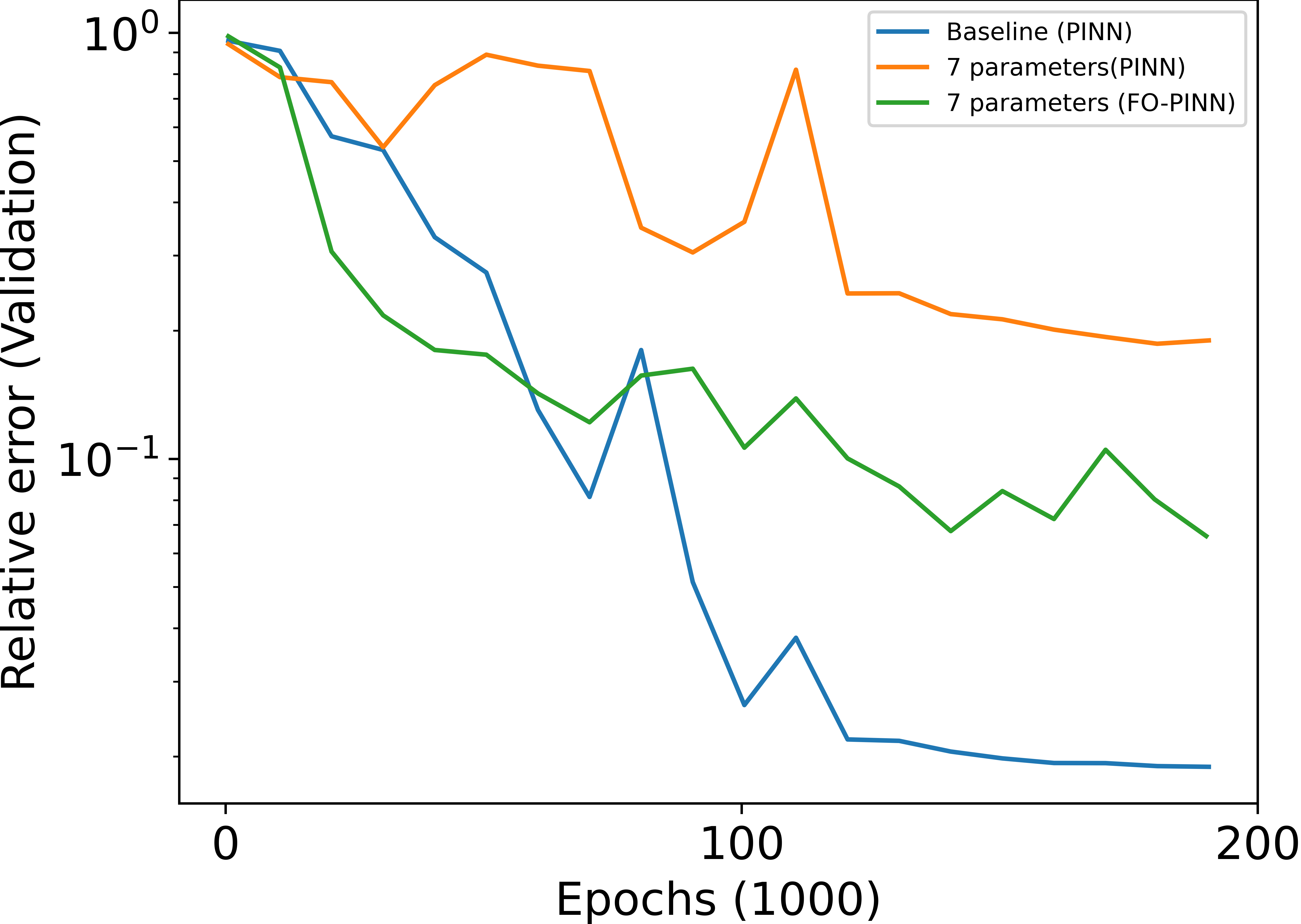}
         \caption{$v$}
         \label{fig:ev}
}
\vspace*{0.2em}
}
\end{subfigure}%
\hfill
\begin{subfigure}[t]{0.32\textwidth} 
\vbox{
\vspace*{0.2em}%
\centering{
         \includegraphics[width=\textwidth]{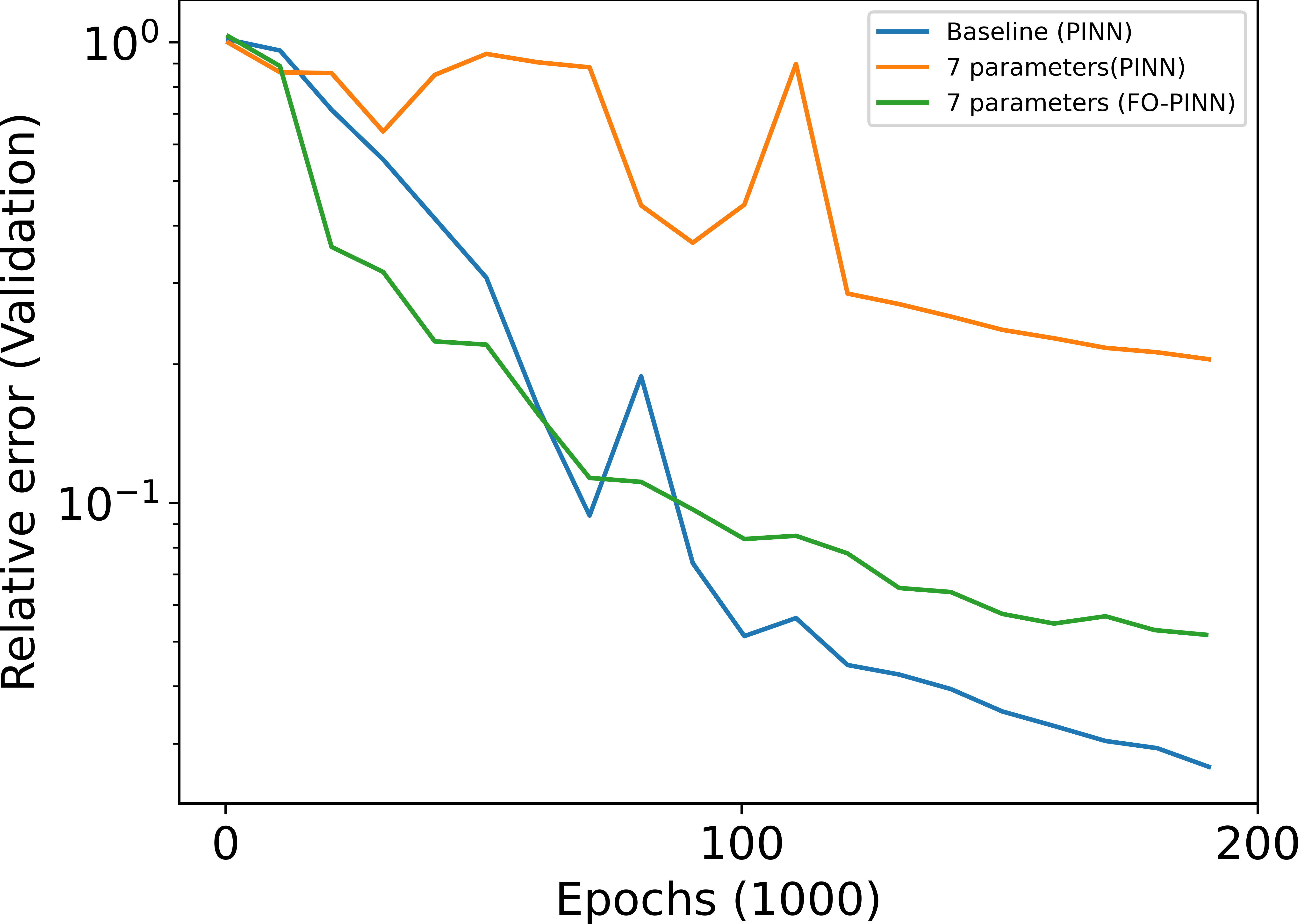}
         \caption{$p$}
         \label{fig:ep}
}
\vspace*{0.2em}
}
\end{subfigure}%
     \caption{Validation error plots for the prediction of flow of in a parameterized system governed by the Navier-Stokes equations.}
        \label{fig:param_cylinder}
\end{figure*}

\begin{figure*}[ht]
     \centering
     \begin{subfigure}[t]{0.4\textwidth}
         \vbox{
\vspace*{0.1em}%
\centering{
         \includegraphics[width=\textwidth]{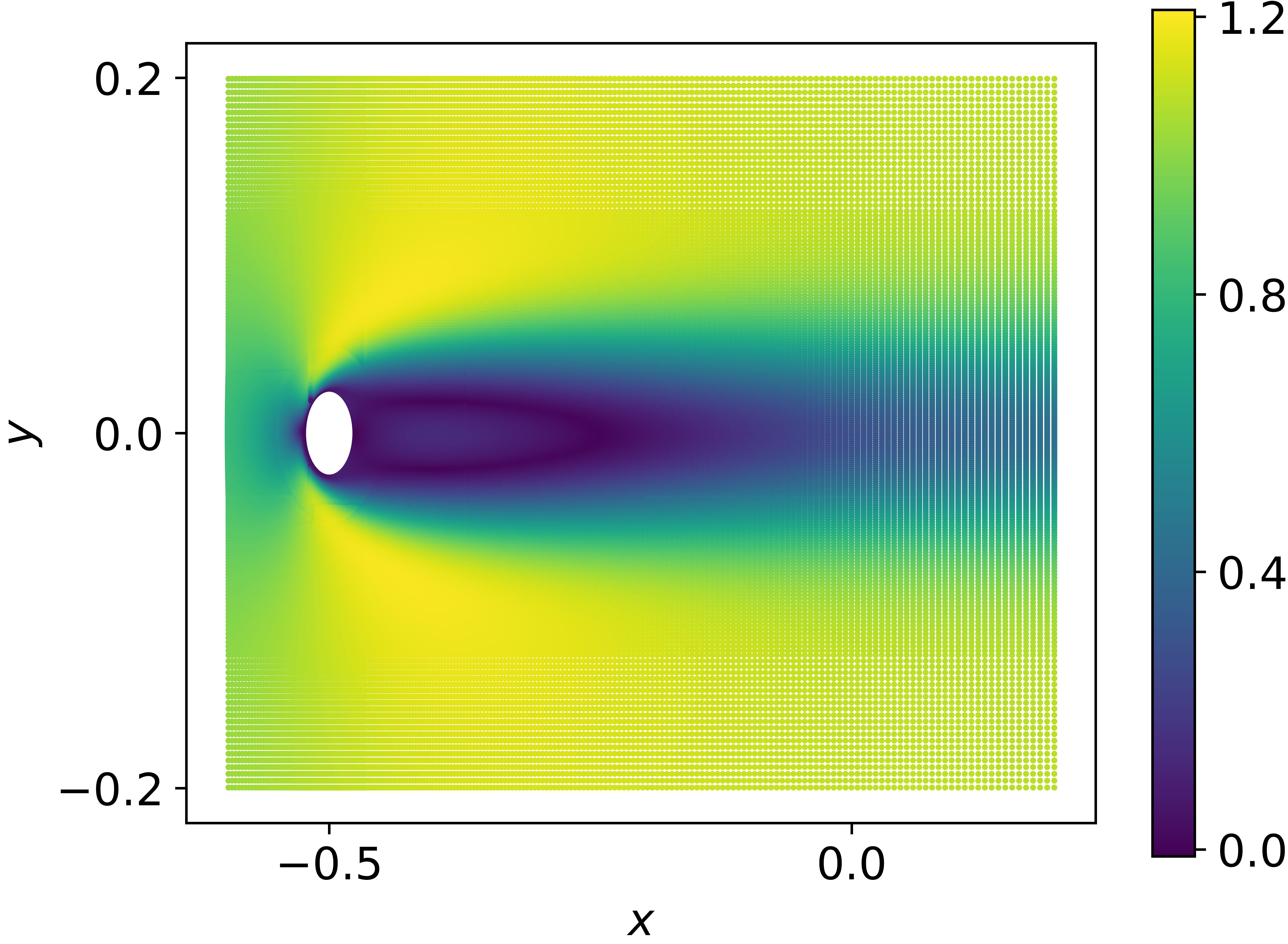}
         \caption{Velocity - PINN}
         \label{fig:annular_uv_pinn}
         }%
\vspace*{0.1em}
}%
     \end{subfigure}
     \hfill
     \begin{subfigure}[t]{0.4\textwidth}
\vbox{
\vspace*{0.1em}%
\centering{
         \includegraphics[width=\textwidth]{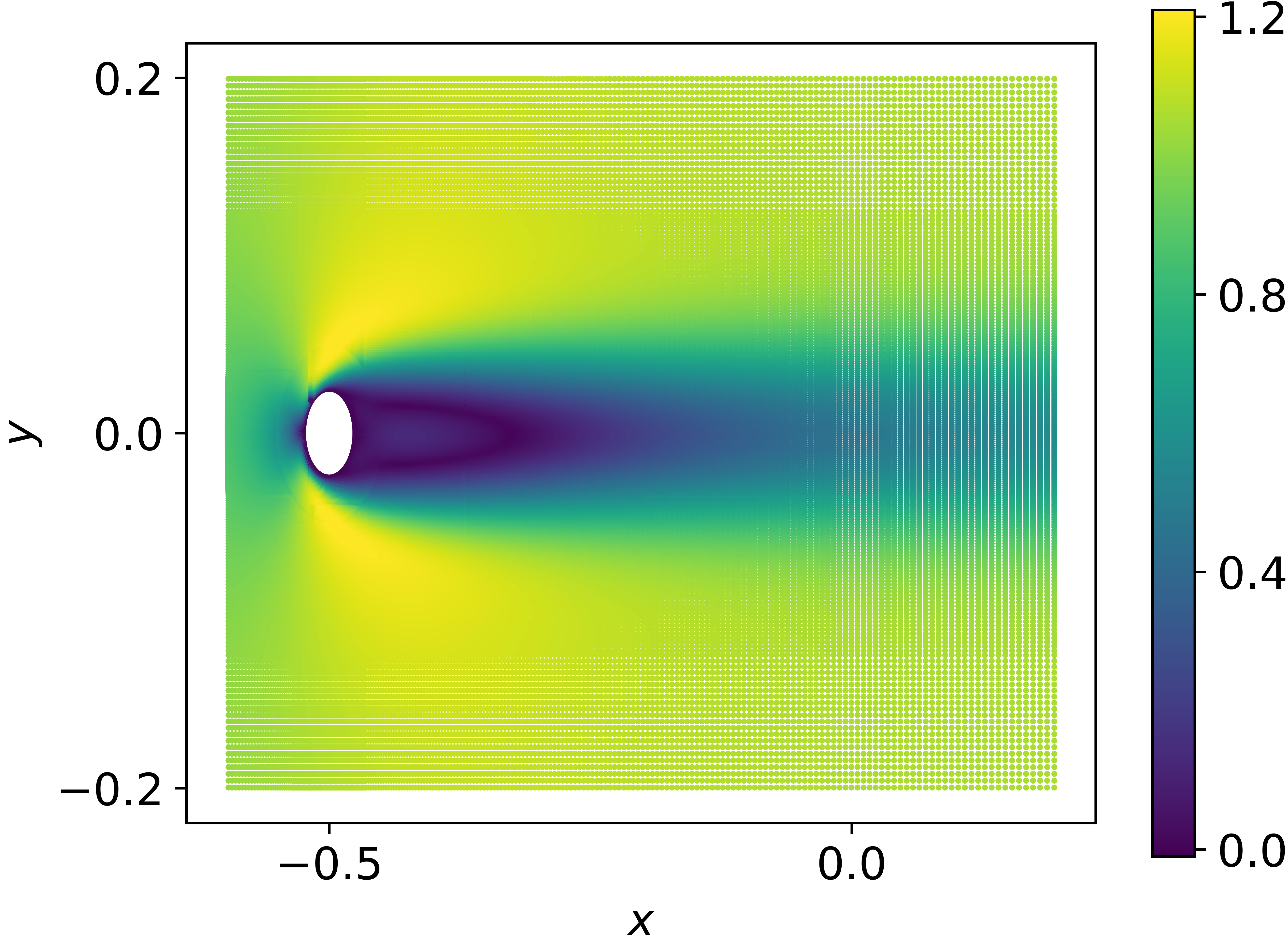}
         \caption{Velocity - FO-PINN}
         \label{fig:annular_uv_fo}
         }%
\vspace*{0.1em}
}%
     \end{subfigure}
     \hfill
     \begin{subfigure}[t]{0.4\textwidth}
\vbox{
\vspace*{0.1em}%
\centering{
         \includegraphics[width=\textwidth]{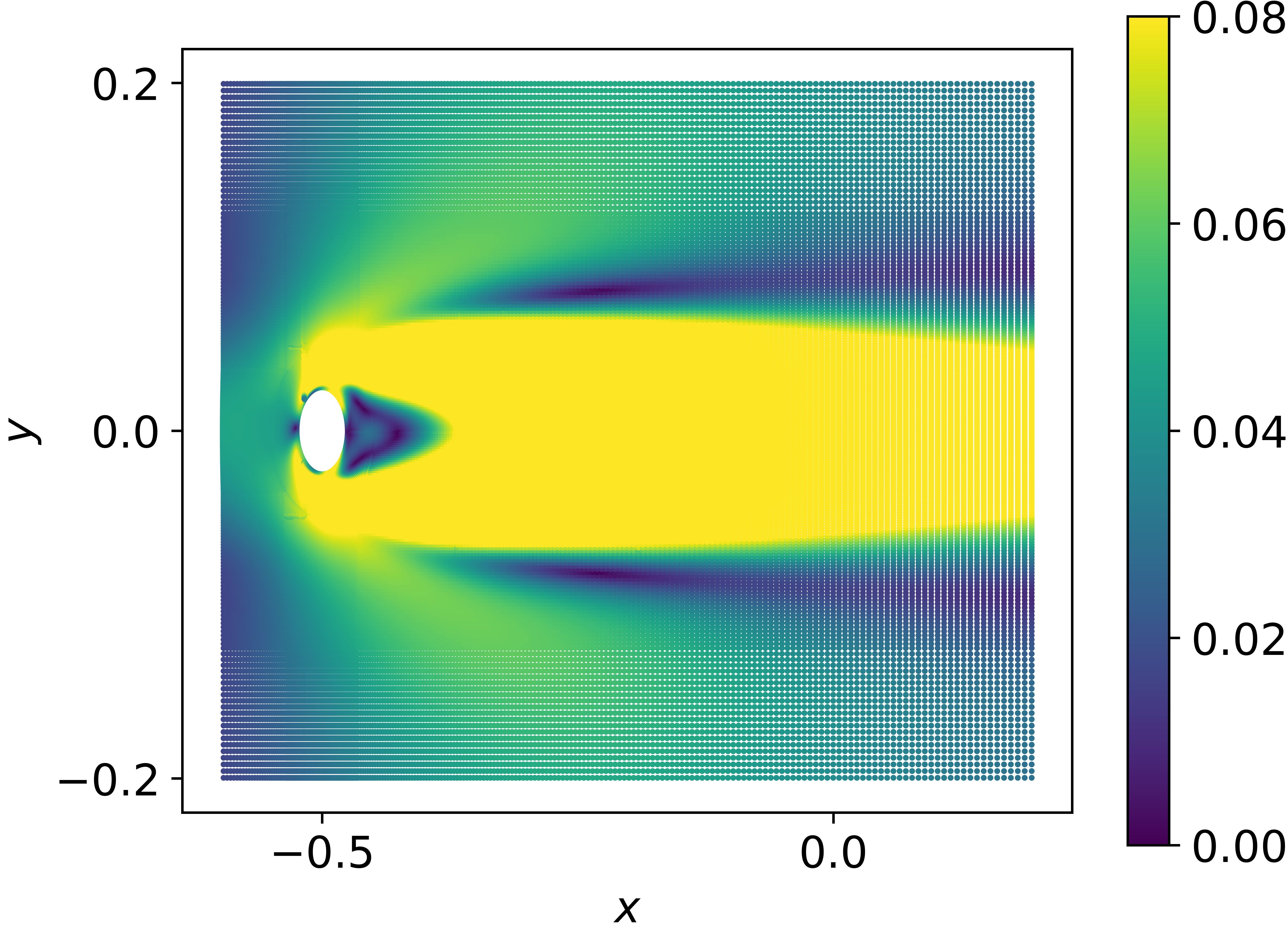}
         \caption{Error for velocity - PINN}
         \label{fig:annular_uv_pinn_diff}
         }%
\vspace*{0.1em}
}%
     \end{subfigure}
     \hfill
     \begin{subfigure}[t]{0.4\textwidth}
\vbox{
\vspace*{0.1em}%
\centering{
         \includegraphics[width=\textwidth]{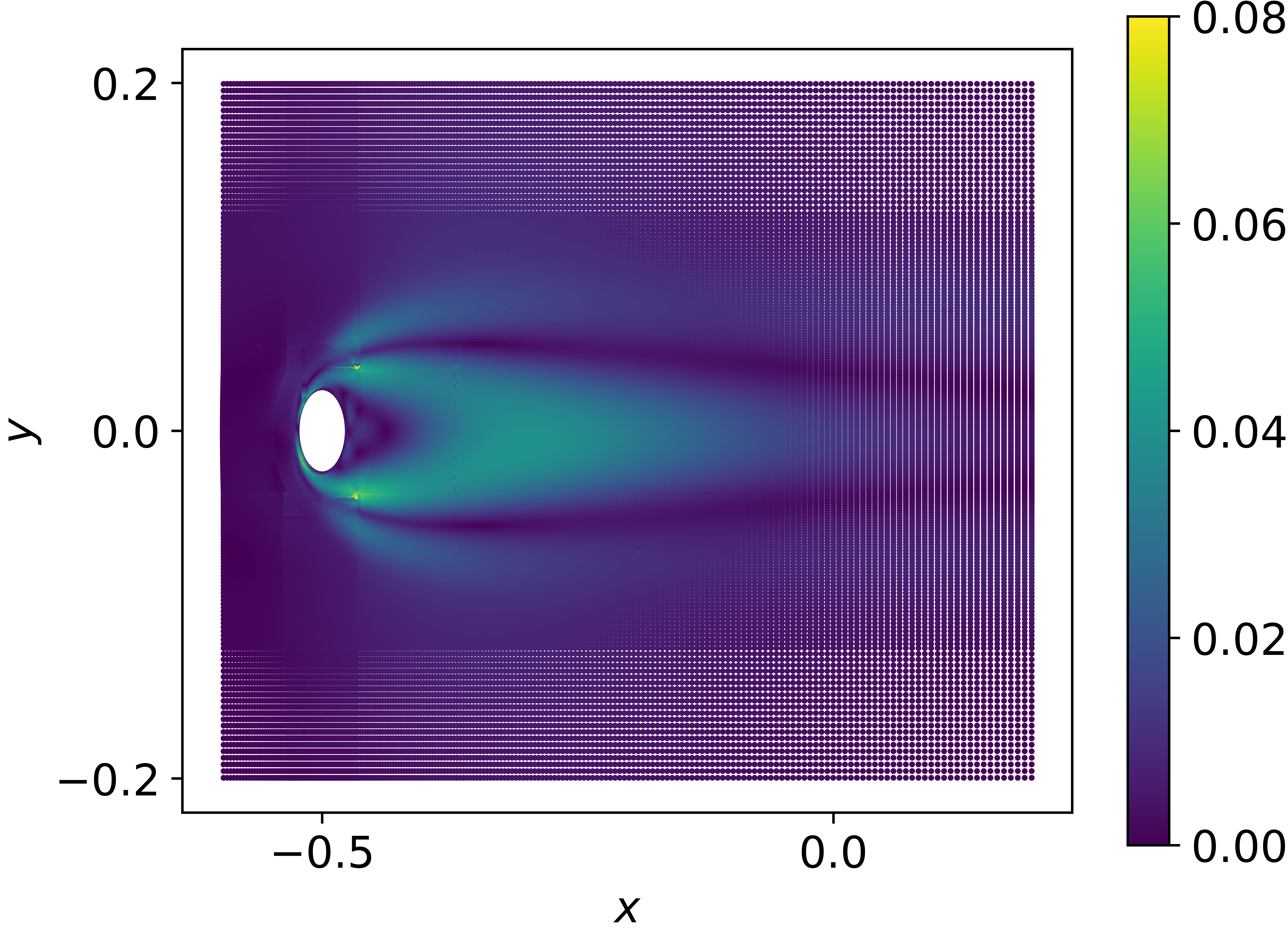}
         \caption{Error for velocity - FO-PINN}
         \label{fig:annular_uv_fo_diff}
         }%
\vspace*{0.1em}
}%
     \end{subfigure}
    \bigskip
     \begin{subfigure}[t]{0.4\textwidth}
\vbox{
\vspace*{0.1em}%
\centering{
         \includegraphics[width=\textwidth]{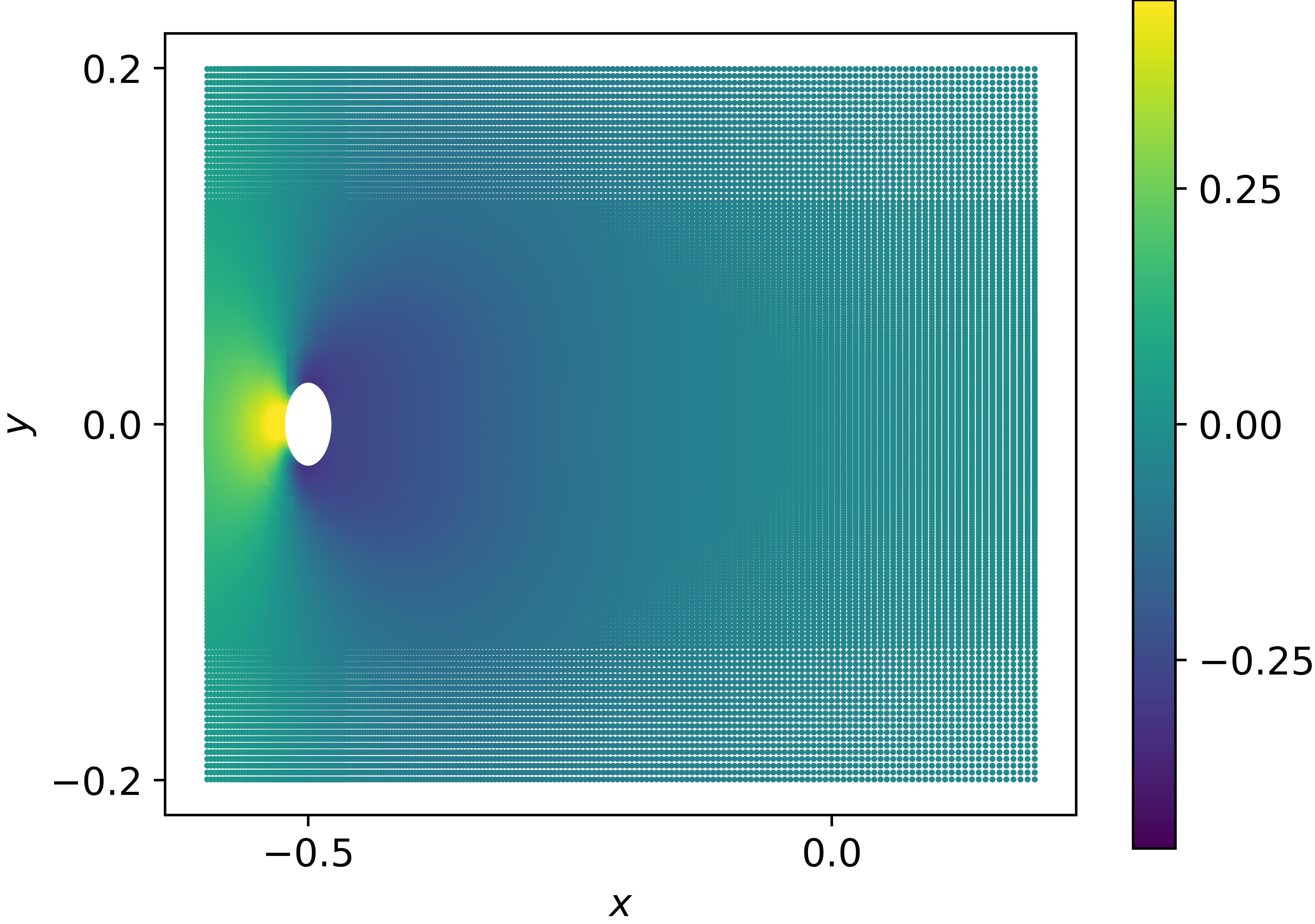}
         \caption{Pressure - PINN}
         \label{fig:annular_p_pinn}
         }%
\vspace*{0.1em}
}%
     \end{subfigure}
     \hfill
     \begin{subfigure}[t]{0.4\textwidth}
\vbox{
\vspace*{0.1em}%
\centering{
         \includegraphics[width=\textwidth]{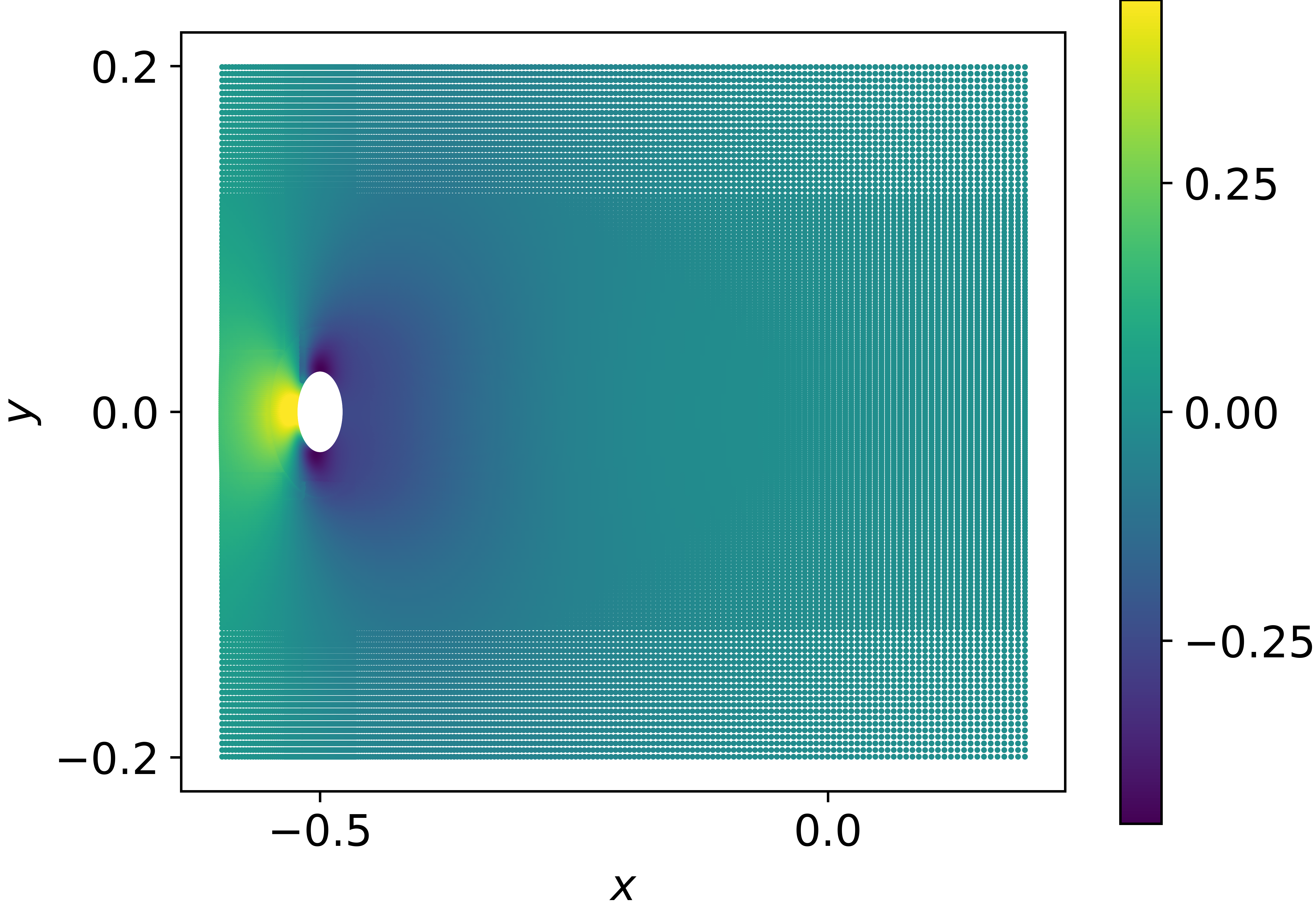}
         \caption{Pressure - FO-PINN}
         \label{fig:annular_p_fo}
         }%
\vspace*{0.1em}
}%
     \end{subfigure}
     \hfill
     \begin{subfigure}[t]{0.4\textwidth}
\vbox{
\vspace*{0.1em}%
\centering{
         \includegraphics[width=\textwidth]{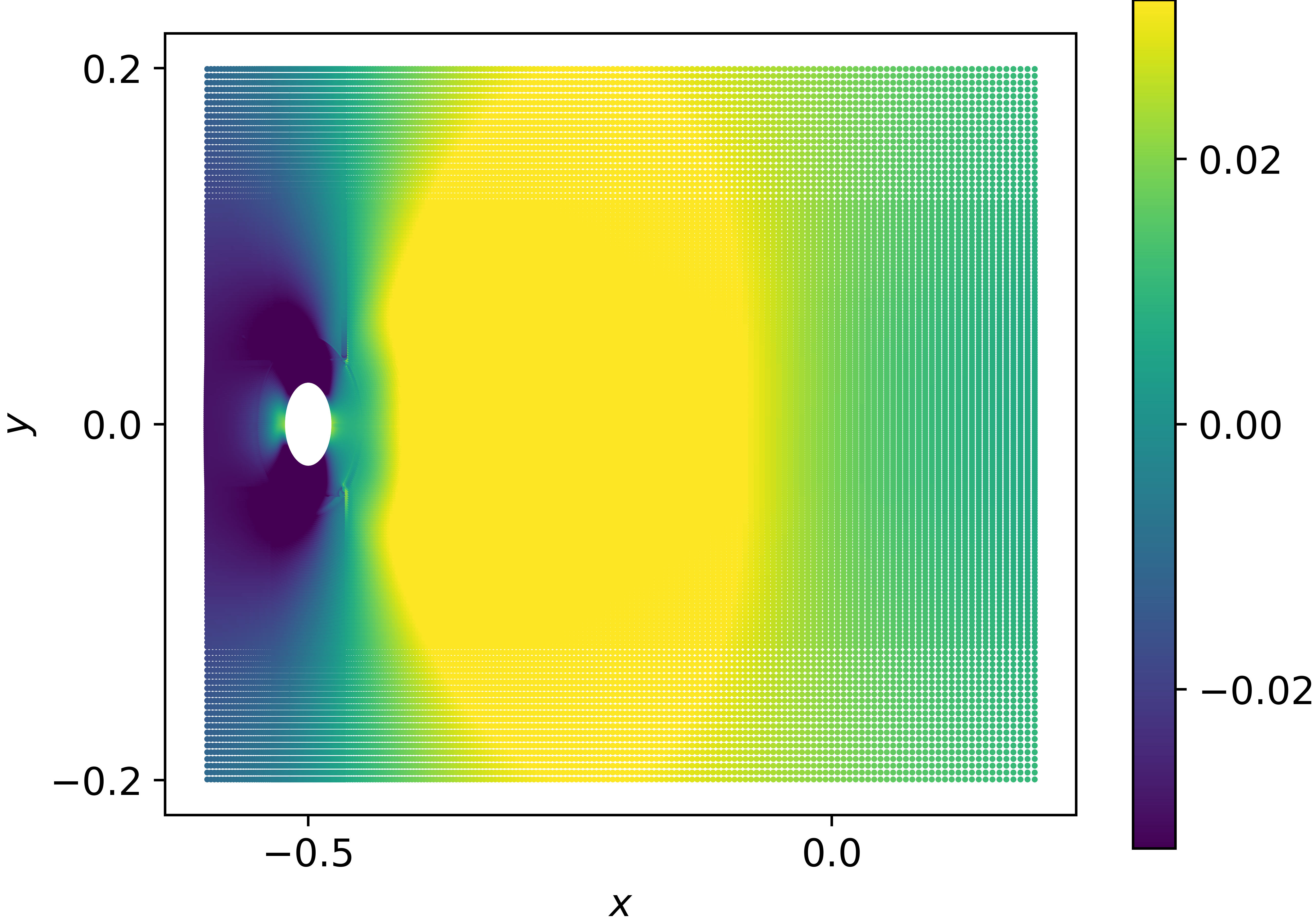}
         \caption{Error for pressure - PINN}
         \label{fig:annular_p_pinn_diff}
         }%
\vspace*{0.1em}
}%
     \end{subfigure}
     \hfill
     \begin{subfigure}[t]{0.4\textwidth}
\vbox{
\vspace*{0.1em}%
\centering{
         \includegraphics[width=\textwidth]{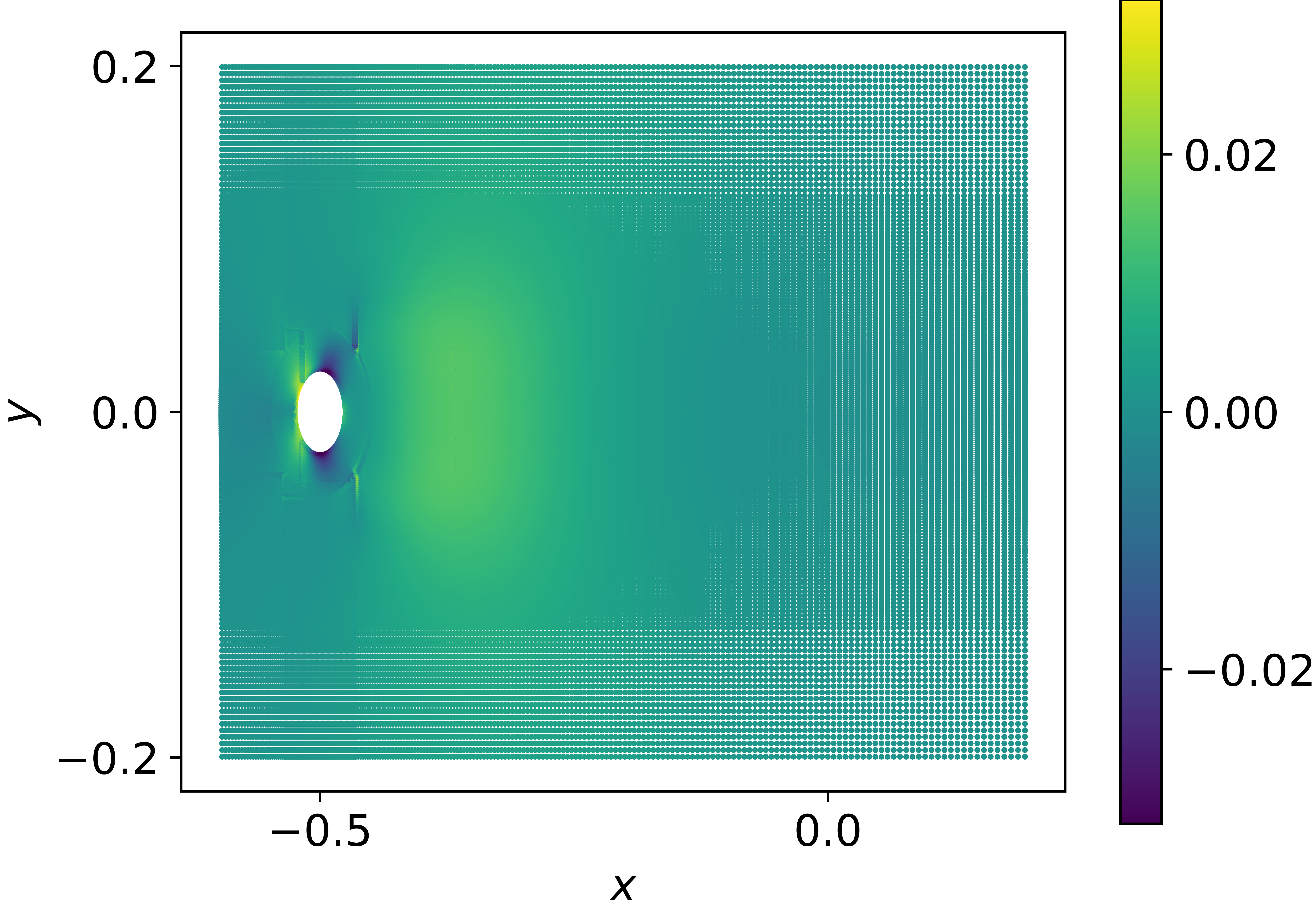}
         \caption{Error for pressure - FO-PINN}
         \label{fig:annular_p_fo_diff}
         }%
\vspace*{0.1em}
}%
     \end{subfigure}
     \caption{Predicted flow on the parameterized system (with 7 parameters) governed by Navier-Stokes equations using PINN and FO-PINN. Figures (a) through (d) show the predicted velocity magnitude from the two networks and their prediction error. Similarly, Figures (e) through (h) show the predicted pressure and the prediction error.}
        \label{fig:annular_ring_p}
\end{figure*}

The velocities in $x$ and $y$ directions and pressure, given by, $u$, $v$, $p$ respectively are the the three variables of interest for this problem. These are the ouputs for a standard PINN. For FO-PINN, however, in addition to these variables, we add  $u_x$, $u_y$, $v_x$, and $v_y$ as outputs of the network similar to the previous example. They represent the first-order spatial derivatives of $u$ and $v$. The first-order formulation of Eq.~\eqref{eq_ns} used for this example is given in Eq.~\eqref{eq_ns_firstorder}. For the predicted outputs of the network $\hat{u}$, $\hat{u}_x$ and $\hat{u}_y$, and $\hat{v}$, $\hat{v}_x$ and $\hat{v}_y$,  additional loss terms are introduced to ensure compatibility over the derivatives as defined in Eq.~\eqref{eq_firstorder_compatibility_ns}. 

Using high-fidelity OpenFOAM simulation results as the "ground truth", Fig. \ref{fig:param_cylinder} compares the validation error between the parameterized FO-PINN model and non-parameterized (baseline) and parameterized PINN models. We observe that the error in FO-PINN are lower by an order of magnitude compared to that of PINN for the parameterized system. The FO-PINN for the parameterized system has accuracies closer to that of the baseline, whereas PINN for the parameterized system performs significantly worse compared to that of its non-parameterized counterpart. This can be also observed clearly in the predicted results and errors for velocity and pressure as shown in Fig. \ref{fig:annular_ring_p}. Figure \ref{fig:param_trend} shows the error trend when the number of parameters are increased. All the models are trained for 200,000 epochs. We can observe that FO-PINN performs stays consistently close to the baseline (zero parameters case), whereas it increases quickly with the number of parameters for standard PINN.

\section{Conclusion}
\label{conclusion}
In this work, we presented FO-PINNs, a first order formulation of Physics-Informed Neural Networks that can be used to solve second and higher-order PDEs with only first-order derivative calculations using automatic differentiation. With numerical examples involving second order PDEs, we showed how FO-PINNs can enable exact BC imposition and training with AMP. In the first example, we showed that FO-PINNs can be trained 2.9x faster compared to PINNs, and results in validation error that is one order of magnitude smaller compared to the PINN model. In the second example, we also showed that FO-PINNs provide significantly more accurate results for parameterized systems compared to PINNs. Although the results we presented are for second order PDE problems, the underlying approach is generalizable to problems that involve higher-order PDEs as well. This will involve only adding new output variables and compatibility equations, which results in a marginal increase in network parameters only in the last layer. To summarize, FO-PINNs offer two main advantages: (1) Increased accuracy for parameterized and non-parameterized problems by smoothing out the sharp variation in second and higher-order derivatives and by imposing BCs exactly, and (2) training speedup by removing extra backpropagation steps for computing the higher-order derivatives, and by using AMP for training. As future work, the performance of FO-PINNs for 3D domains and problems with higher-order PDEs will be investigated.

One of the biggest promises of the PINNs in industrial applications is the capability of solving for parameterized systems in a single training, whereas traditional solvers are limited to non-parameterized simulations. These parameterized models can be used in developing industrial digital twins with real-time predictions. However, standard PINNs suffer from accuracy decline as the dimensionality of the parameter space increases. FO-PINNs are a viable approach to address this accuracy decline, and take us one step closer to developing reliable AI solutions for industrial systems and facilitating scientific computing. Moreover, the training speedup offered by FO-PINNs opens the door to accelerated industrial design procedures and enables developing more expressive physics-informed models and more optimal hyperparameter tuning.
\section*{Acknowledgment} 

Hadi Meidani and Rini Gladstone acknowledge the support by the National Science Foundation under Grant No. CMMI-1752302.

\printbibliography

\end{document}